\documentclass[lettersize,journal]{IEEEtran}
\usepackage{amsmath,amsfonts}
\usepackage{algorithmic}
\usepackage{algorithm}
\usepackage{array}
\usepackage[caption=false,font=normalsize,labelfont=sf,textfont=sf]{subfig}
\usepackage{textcomp}
\usepackage{stfloats}
\usepackage{url}
\usepackage{verbatim}
\usepackage{graphicx}
\usepackage{soul}
\soulregister{\cite}7 
\soulregister{\citep}7 
\soulregister{\citet}7 
\soulregister{\ref}7 
\soulregister{\pageref}7 

\usepackage{multirow}
\usepackage{makecell}
\usepackage{pifont}
\usepackage{booktabs} 

\usepackage{xcolor}
\definecolor{darkred}{rgb}{0.99, 0.2, 0.5}
\definecolor{darkyellow}{rgb}{0.99, 0.79, 0.19}
\usepackage{colortbl}
\usepackage{balance}

\usepackage{cite}
\begin{document}

\title{Crucial-Diff: A Unified Diffusion Model for Crucial Image and Annotation Synthesis in Data-scarce Scenarios}

\author{Siyue Yao, Mingjie Sun\textsuperscript{\dag},~\IEEEmembership{Member,~IEEE}, Eng Gee Lim,~\IEEEmembership{Senior Member,~IEEE}, Ran Yi,~\IEEEmembership{Member,~IEEE}, \\ Baojiang Zhong,~\IEEEmembership{Senior Member,~IEEE}, Moncef Gabbouj,~\IEEEmembership{Fellow,~IEEE}
\thanks{\dag{ }{Corresponding Author.}}
\thanks{This work was supported by Young Scientists Fund of the National Natural Science Foundation of China (Grant No. 62302328), Jiangsu Province Foundation for Young Scientists (Grant No. BK20230482), Suzhou Key Laboratory Open Project (Grant No. 25SZZD07) and Jiangsu Manufacturing Strong Province Construction Special Fund Project (Grant Name: Research and Development and Industrialization of Intelligent Service Robots Integrating Large Model and Multimodal Technology).}
\thanks{Siyue Yao, and Eng Gee Lim are with the School of Advanced Technology, Xi'an Jiaotong-Liverpool University, Suzhou 215123, China (e-mail: siyue.yao2302@student.xjtlu.edu.cn; enggee.lim@xjtlu.edu.cn).}
\thanks{Mingjie Sun, and Baojiang Zhong are with the School of Computer Science and Technology, Soochow University, Suzhou 215006, China (e-mail: mjsun@suda.edu.cn; bjzhong@suda.edu.cn).}
\thanks{Ran Yi is with the School of Computer Science, Shanghai Jiao Tong University, Shanghai 200240, China (e-mail: ranyi@sjtu.edu.cn).}
\thanks{Moncef Gabbouj is with the Faculty of Information Technology and Communication Sciences, Tampere University, Tampere 33720, Finland (e-mail: moncef.gabbouj@tuni.fi).}
}

\markboth{IEEE TRANSACTIONS ON IMAGE PROCESSING}%
{Shell \MakeLowercase{\textit{et al.}}: A Sample Article Using IEEEtran.cls for IEEE Journals}


\maketitle

\begin{abstract}
The scarcity of data in various scenarios, such as medical, industry and autonomous driving, leads to model overfitting and dataset imbalance, thus hindering effective detection and segmentation performance. 
Existing studies employ the generative models to synthesize more training samples to mitigate data scarcity. 
However, these synthetic samples are repetitive or simplistic and fail to provide ``crucial information" that targets the downstream model's weaknesses. Additionally, these methods typically require separate training for different objects, leading to computational inefficiencies.
To address these issues, we propose Crucial-Diff, a domain-agnostic framework designed to synthesize crucial samples. Our method integrates two key modules. 
The Scene Agnostic Feature Extractor (SAFE) utilizes a unified feature extractor to capture target information. The Weakness Aware Sample Miner (WASM) generates hard-to-detect samples using feedback from the detection results of downstream model, which is then fused with the output of SAFE module.
Together, our Crucial-Diff framework generates diverse, high-quality training data, achieving a pixel-level AP of 83.63\% and an F1-MAX of 78.12\% on MVTec. On polyp dataset, Crucial-Diff reaches an mIoU of 81.64\% and an mDice of 87.69\%. Code is publicly available at \url{https://github.com/JJessicaYao/Crucial-diff}.
\end{abstract}


\section{Introduction}

\begin{figure*}[t!]
    \centering
    \includegraphics[width=1\linewidth]{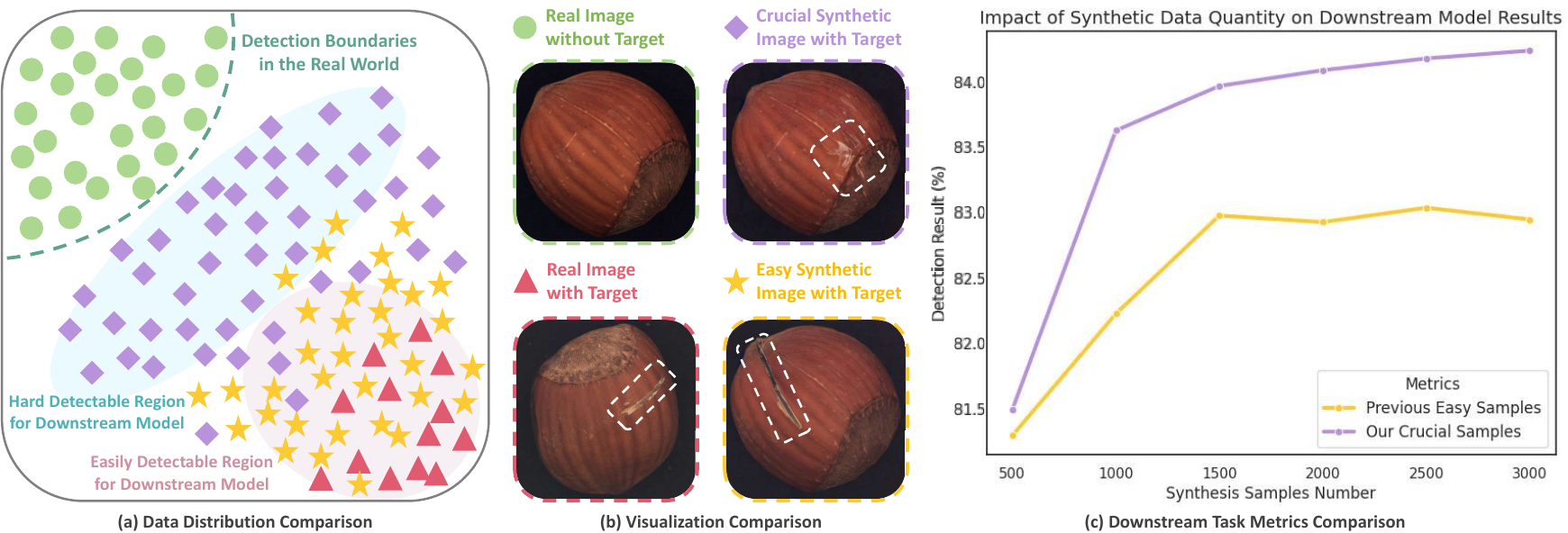}
    \caption{Comparison of previous synthesis methods and our Crucial-Diff. (a) In data distribution comparison, our method generates images \textcolor{violet}{(purple points)} that are more challenging for downstream tasks, while previous methods produce samples \textcolor{darkyellow}{(yellow points)} that closely resemble real images \textcolor{darkred}{(pink points)} and easy detection. (b) The visualization comparison shows that our samples are less identifiable, while prior methods replicate target patterns from the original dataset, with white dashed boxes indicating target areas. (c) The downstream task metrics comparison highlights the pixel-level performance of a U-Net model trained on synthetic images. As data volume increases, our method consistently improves results, while previous methods decline, emphasizing the importance of generating crucial data.}
    \label{fig:intro}
\end{figure*}

The scarcity of training samples poses significant challenges in developing reliable deep-learning models, such as the scenarios of the industrial quality inspection with rare defect images \cite{roth2022towards,yao2023explicit,you2022unified,liang2023omni,gu2023remembering,liu2024cross,li2024promptad} and the medical image diagnosis with rare precise annotations \cite{zhang2022lesion, lu2024anomaly,liu2024multimodal,li2024single,qiu2024learn,li2025srconvnet,cong2025reference}. Therefore, this study focuses on target segmentation tasks in these data-scarce situations, aiming to investigate methods for generating effective samples to alleviate data insufficiency.
To address this challenge, the Images and Annotations Synthesis (IAS) task in data-scarce scenarios has emerged as a pivotal paradigm \cite{hu2023anomalydiffusion, ControlPolypNet,Guanfewshotgeneration,du2023arsdm,tan2021night}. It synthesizes diverse image samples with accurate segmentation annotations to expand limited data, enabling large-scale synthetic data to train and improve downstream models.

Early IAS methods in data-scarce scenarios relied on traditional data augmentation techniques \cite{li2021cutpaste}, expanding the dataset through transformations like rotation and cropping, \emph{etc}. In contrast, modern generative approaches, including Generative Adversarial Networks (GANs) \cite{niu2020defect, zhang2021defect} and diffusion models \cite{hu2023anomalydiffusion, Guanfewshotgeneration}, aim to replicate training data distributions by minimizing statistical divergence between synthetic and real samples.
As shown in Fig. \ref{fig:intro}(a), samples generated by these methods, which we refer to as ``easy synthetic samples" \textcolor{darkyellow}{(yellow points)}, cluster densely around real samples \textcolor{darkred}{(pink points)}. 
This limitation is further highlighted in Fig. \ref{fig:intro}(b) of the industrial inspection scenario, where easy synthetic images appear visually identical to real anomalies, failing to introduce meaningful variations or challenges for downstream models.

As shown in the  \textcolor{darkyellow}{yellow curve} of Fig. \ref{fig:intro}(c), only using the easy synthetic samples (the \textcolor{darkyellow}{yellow points} in Fig. \ref{fig:intro}(a)) initially boosts downstream task performance.
However, this improvement quickly plateaus and may even decline as the data volume increases. 
This stagnation stems from the tendency of easy synthetic samples to \textbf{cluster around simple, known patterns}. While these samples quantitatively expand the dataset, they lack qualitative novelty or innovative features, causing downstream models to overfit on familiar patterns and exhibit poor generalization in unseen scenarios.
In contrast, ``crucial synthetic images" specifically designed to target uncertain and hard detectable feature of the downstream models, as shown by the \textcolor{violet}{purple points} in Fig. \ref{fig:intro}(a). These samples expose the weaknesses of downstream models, significantly improving their accuracy by forcing the models to learn from these challenging samples.

Additionally, existing approaches such as DFMGAN \cite{duan2023few} often train separate generative models for each target due to significant domain gaps between different targets. This per target training paradigm \textbf{consumes substantial computational resources}, resulting in exponentially increasing costs as the number of required targets grows. Such methods severely constrain the scalability of synthetic data, ultimately limiting their widespread adoption in real world applications.

Given these limitations, a question arises: \textbf{\textit{What kind of synthetic data is more effective in enhancing downstream model performance?}} The answer lies in addressing two core principles: (1) generating crucial synthetic samples that challenge downstream models by incorporating ambiguous textures or deceptive characteristics, \emph{etc}; and (2) ensuring the synthetic data is easily acquired, which requires a universally applicable pipeline that generates high-quality data across diverse domains without domain-specific redesign or retraining. By focusing on these principles, synthetic data can effectively target the weaknesses of downstream models, driving continuous improvements in robustness and generalization while minimizing training resource consumption through a domain-agnostic framework.


To address these requirements, we propose \textbf{Crucial-Diff}, a domain-agnostic framework that focuses on crucial synthetic sample generation. 
By employing a diffusion model conditioned on the reference image of the target, Crucial-Diff synthesizes realistic targets on a background image without any target, positioned according to the specified box location, thereby eliminating the fine-tuning needs for different targets or scenarios. 
In contrast to previous methods dependent on precise masks as location constraints for synthesis, which typically require additional mask generators,  our method randomly generates bounding box masks without needing extra generators, significantly reducing computational overhead.
Furthermore, two additional modules are proposed to address crucial synthetic sample generation and multi-scenario universality, respectively, to ensure that the generated samples are effectively bridge data scarcity.

To meet the requirement (1), we introduced the Weakness Aware Sample Miner (WASM) module, specifically designed to identify the inherent weaknesses of downstream models, particularly regarding frequently misclassified images. The WASM module employs a feedback loss that minimizes the downstream model detection difference between the target and background regions, making it challenging to distinguish between them. In this way, the generated crucial synthetic images often exhibit ambiguous textures, inconspicuous targets or other deceptive characteristics, as shown in \textcolor{violet}{purple part} of Fig. \ref{fig:intro}(b). When used to train downstream models, these samples significantly improve robustness, enabling sustained performance gains without overfitting, as illustrated by the \textcolor{violet}{purple curve} in Fig. \ref{fig:intro}(c).

To meet the requirement (2), our framework introduces a Scene Agnostic Feature Extractor (SAFE) module based on the pre-trained visual encoder \cite{radford2021learning}, enabling a single real target image to serve as a reference. SAFE transforms these visual references into textual descriptions of targets, which are then fed into the diffusion model for synthesizing precise targets. By simply modifying the reference image, diverse targets can be generated without redesign or retraining, significantly reducing training resource consumption. 

By harmoniously integrating two proposed modules with the diffusion model, our method, Crucial-Diff, effectively addresses the IAS task and provides high-quality crucial training data, which significantly improves subsequent detection and segmentation tasks.
Our contributions can be encapsulated within the following three aspects:
\begin{itemize}
    \item 
    We introduce Crucial-Diff, a domain-agnostic framework to tackle data scarcity in downstream tasks by generating high-quality samples. Therefore, the Weakness Aware Sample Miner (WASM) module creates crucial samples containing deceptive features, which assist downstream models in learning challenging features to improve the accuracy and robustness of downstream models.
    \item  
    To ensure universal applicability, we introduce Scene Agnostic Feature Extractor (SAFE) module, which transforms the visual features of reference images into textual descriptions. This allows for synthesising diverse targets without domain-specific retraining, notably reducing resource costs and making generated data easily accessible.
    \item  
    Extensive experiments indicate our crucial samples effectively improve downstream tasks. On MVTec, AP and F1-MAX increase by 1.4\% and 1.1\% compared to AnomalyDiffusion \cite{hu2023anomalydiffusion}. On polyp dataset, mIoU and mDice raise by 5.3\% and 6.5\% compared to ArSDM \cite{du2023arsdm}.
    
\end{itemize}

\section{Related Work}
\subsection{Text-to-Image Generation}
Text-to-image generation rapidly evolves, primarily driven by advancements in Generative Adversarial Networks (GANs) \cite{goodfellow2014generative} and Variational Auto-Encoders (VAEs) \cite{kingma2013auto}. Early GAN-based models \cite{zhang2017stackgan, xu2018attngan} use attention mechanisms for high-resolution synthesis but struggle with diverse scenes. 
VAE-based models \cite{ding2021cogview, ramesh2021zero} also contribute to the field by providing a probabilistic framework for image generation, though they frequently produce blurry outputs due to reliance on mean-squared error loss.
Recent developments in diffusion models \cite{ho2020denoising, song2019generative} refine noisy latent vectors into desired images through a denoising process. These models, exemplified by Stable Diffusion (SD) \cite{rombach2022high} and DALLE \cite{ramesh2022hierarchical}, offer high-quality outputs, setting new benchmarks in text-to-image synthesis. 
Beyond generic text conditioning, \cite{gal2023image} and \cite{kumari2023multi} propose adaptive word embedding learning from few-shot image examples, overcoming the limitations of standard prompts in capturing fine-grained or user-specific concepts.

\subsection{Rare Image and Annotation Synthesis}
Traditional approaches typically employ GANs to generate image-annotation pairs \cite{kar2019meta,choi2019self}, while contemporary diffusion models demonstrate superior capabilities for image-annotation synthesis \cite{wu2023datasetdm, wu2023diffumask,yi2023towards,feng2024instagen}. DSMA \cite{dai2024diffusion} further enables adversarial attacks using synthetic dataset. However, these methods fail in extreme few-shot scenarios, particularly in industrial and medical applications. Therefore, we focus on developing generation models for these two fields.

\textbf{Industrial inspection.} Early non-generative methods \cite{zavrtanik2021draem, li2021cutpaste, lin2021few} utilize traditional data augmentation to create anomaly images, but these often result in low consistency and fidelity. Inspired by GANs, researchers investigate techniques like SDGAN \cite{niu2020defect} and Defect-GAN \cite{zhang2021defect}, which rely on defect-free samples. DFMGAN \cite{duan2023few} generates anomaly images and masks but struggles with complex objects. Both AnomalyDiffusion \cite{hu2023anomalydiffusion} and AnoGen \cite{Guanfewshotgeneration} are recent approaches based on diffusion models. AnomalyDiffusion requires separate models for accurate mask generation, while AnoGen replaces mask generation with bounding boxes but lacks precise annotations.

\textbf{Medical Diagnosis.} GAN-based methods \cite{shin2018abnormal, he2021colonoscopic, sams2022gan} effectively address the challenge of limited labeled data in the context of colon polyps. For example, DCGAN \cite{sasmal2020improved} enhances polyp classification but often suffers from convergence issues, producing less realistic and diverse images. In contrast, diffusion models effectively address the realism challenge \cite{du2023arsdm, machavcek2023mask} by leveraging latent diffusion models to generate synthetic polyps from segmentation masks. ControlPolypNet \cite{ControlPolypNet} and Siamese-Diffusion \cite{qiu2025noise} enhances realism by precisely controlling backgrounds, significantly improving image quality.
\textit{}

\section{Method}
\begin{figure*}[t]
    \centering
    \includegraphics[width=1\linewidth]{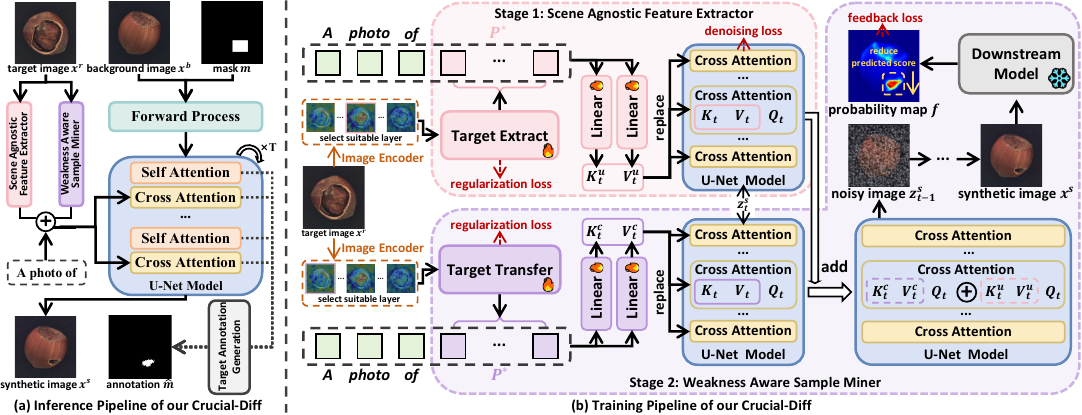}
    \caption{Overview of Crucial-Diff framework. (a) Inference pipeline of Crucial-Diff, which utilizes a mask and a background image as inputs, with a target image to guide the synthesis process. Crucial-Diff generates critical samples and extracts annotations from attention layers. (b) Training pipeline of Crucial-Diff, which consists of two stages. Stage 1 trains a Scene Agnostic Feature Extractor (SAFE) module to convert target images into textual features. Stage 2 trains a Weakness Aware Sample Miner (WASM) module to extract features that challenge downstream models in detecting targets from background, then fuse extracted features with SAFE in cross-attention layer.}
  \label{fig:pipe}
\end{figure*}

Crucial-Diff tackles target synthesis challenges in data-scarce scenarios by generating high-quality samples that expose downstream model weaknesses and ensure cross-domain applicability. 
As shown in Fig. \ref{fig:pipe}(a), our Crucial-Diff synthesizes crucial images by inputting a target reference image $x^r$ that encodes target-specific features, a background image $x^b$ that provides contextual information and a bounding box mask $m$ that specifies the location of target into model. The bounding box mask is generated by randomly sampling coordinates within the background image space, with both its position and size determined through a randomization process.
Crucial-Diff begins with the Scene Agnostic Feature Extractor (SAFE) module, which extracts textual embeddings of the target characteristic from the reference image (Sec. \ref{sec:uni}). 
Next, the Weakness Aware Sample Miner (WASM) extracts textual embeddings of subtle target characteristics that can mislead downstream model detection errors from the reference image, leveraging its ability to capture deceptive features acquired during the training process (Sec. \ref{sec:hard}).
The textual features from WASM module are used to compute attention outputs, which are then fused with those from SAFE module, dynamically guiding synthesis process to obtain crucial images with target. 
Finally, accurate pixel-level annotations of targets are produced by extracting and binarizing attention maps between crucial features and target features derived from proposed two modules (Sec. \ref{sec:anno}). 

\subsection{Basic Pipeline}
Our method leverages the state-of-the-art (SOTA) generative model, SD \cite{rombach2022high}, to address IAS tasks in data scarcity scenarios by inpainting targets at specified locations in background images.
Specifically, during training, SD begins by encoding the real image containing the target $x^r$ into a latent representation using a pre-trained VAE model $\varepsilon(\cdot)$. This yields a latent vector $\nu_0^r = \varepsilon(x^r)$ with dimensions $W \times H \times D$, where $W$ and $H$ denote the image size, and $D$ is the number of channels. Following this, Gaussian noise is incrementally added to $\nu_0^r$ as follows:

\begin{equation}
\nu_t^r=\sqrt{\bar{\alpha}_t}\nu_0^r+\sqrt{1-\bar{\alpha}_t}\mathcal{I} ,
\label{equ:forward}
\end{equation}
where $\nu_t^r$ is the noisy latent vector, and $\alpha_t$ is the standard deviation of the noise at timestep $t$. $\bar{\alpha}_t$ represents the cumulative value from $\alpha_0$ to $\alpha_t$. $\mathcal{I}$ denotes the identity matrix.

At each timestep $t$, the noisy latent vector $\nu^r_t$, the bounding box mask $m$, and the masked image latent $\varepsilon(x^r(1-m))$ are concatenated to form a composite feature $z^r_t$.
Meanwhile, to guide the synthesis, a text encoder processes input prompts into embeddings $p \in \mathbb{R}^{l \times q}$, where $l$ is the number of tokens and $q$ is the embedding dimensionality. Finally, a U-Net model is then utilized to predict the noise at each step, aiming to minimize the discrepancy between its output and the actual noise added. This training objective ensures that the model learns to accurately reconstruct targets:

\begin{equation}
\mathcal{L} = \mathbb{E}_{z_t^r,p,t,\epsilon}\big[\big|\big|\epsilon - \epsilon_\theta(z_t^r, p, t)\big|\big|^2\big].
\label{equ:loss}
\end{equation}

During inference, SD encodes a composite feature $z_T^s$, 
which consists of random noise, the masked feature of an arbitrary background image (without the target) and a bounding box mask, similar to training phase. Afterward, $z_T^s$ will be progressively refined into a clean latent vector $z^s$ that incorporates the desired target, using the noise predicted by the well-trained U-Net model, which can be formulated as:

\begin{equation}
z_{t-1}^s=\frac1{\sqrt{\alpha_t}}\left(z_t^s-\frac{1-\alpha_t}{\sqrt{1-\bar{\alpha}_t}}\epsilon_\theta \Big(z_t^s, p, t\Big) \right)+\sqrt{1-\alpha_t} \mathcal{I}  ,
\label{equ:reverse}
\end{equation}
where $\epsilon_\theta (z_t^s, p, t)$ is the model predicted noise at timestep $t$.

To incorporate text information in the generation process, SD utilizes a cross-attention mechanism that operates on three inputs: a query vector $\mathbf{Q}$, a key vector $\mathbf{K}$, and a value vector $\mathbf{V}$. Specifically, at any timestep $t$, $\mathbf{Q}_t$ is obtained from a linear projection of the feature maps produced by each convolutional block, while $\mathbf{K}_t=(\mathcal{W}_t)^K\cdot p$ and $\mathbf{V}_t=(\mathcal{W}_t)^V \cdot p$ are computed through linear projections of the textual embeddings $p$. The cross-attention mechanism is computed in two steps. First, the similarity between $\mathbf{Q}_t$ and $\mathbf{K}_t$ is measured, followed by a softmax operation to normalize the attention weights. These attention weights are then applied to value vector $\mathbf{V}_t$, which is expressed as:
\begin{equation}
Attn(\mathbf{Q}_t, \mathbf{K}_t, \mathbf{V}_t) = \mathrm{softmax}(\frac{\mathbf{Q}_{t}^{}\mathbf{K}_{t}^\top{}}{\sqrt{d}}) \mathbf{V}_t,
\label{equ:att}
\end{equation}
where $\sqrt{d}$ is a scaling factor. This mechanism ensures that the model dynamically focuses on relevant regions of the image while incorporating semantic guidance from the text.

\subsection{Scene Agnostic Feature Extractor Module}
\label{sec:uni}
Unlike previous approaches requiring fine-tuning separate models for different targets, our method introduces a unified framework that eliminates this inefficiency.
To achieve this, we propose the Scene Agnostic Feature Extractor (SAFE) module, which addresses the limitations of existing methods by enabling cross-domain generalization without the need for per-target fine-tuning. The SAFE module leverages a pre-trained image encoder to extract hierarchical visual features from a reference image with the target. These features are then mapped into textual embeddings, which guide SD to generate synthetic data across diverse domains.

In stage 1 of Fig. \ref{fig:pipe}(b), given a real scarce image $x^r$, serving as a concept reference of the target, we first extract its visual features using the CLIP visual encoder $\Psi(\cdot)$ following \cite{zhou2023anomalyclip,gu2024filo}. To capture both low-level and high-level semantics, we retain features from $N$ intermediate layers of the encoder like \cite{wei2023elite}, since only specific layers contain meaningful information about target. Each layer's feature $\Psi_n(x^r)$ is then projected into textual embedding space via a target extractor $\phi^u(\cdot)$:

\begin{equation}
p_n^r = \phi^u(\Psi_n(x^r)), \quad p_n^r \in \mathbb{R}^{1 \times q},
\label{equ:cond}
\end{equation}
where $n \in \{1, 2, \ldots, N\}$ denotes the layer number of the CLIP visual encoder, with a total of $N$ layers. To reduce redundant computations, only $\Gamma$ textual embedding is selected and concatenated along the token dimension to generate a composite textual embedding $p^*\in \mathbb{R}^{\Gamma \times q}$. Besides, to enhance semantic flexibility, $p^*$ is fused with textual embedding of base prompt templates, such as ``a photo of \{$p^*$\}", yielding the final embedding $p^{u}$ to guidance generation. 

Finally, the embedding $p^{u}$ is projected into key and value vectors using two learnable linear blocks, $(\mathcal{W}_t^u)^K$ and $(\mathcal{W}_t^u)^V$. These projected key $\mathbf{K}_t^u$ and value $\mathbf{V}_t^u$ replace the original key and value vectors in the cross-attention mechanism of Eq. \ref{equ:att}, yielding the following updated output:

\begin{equation}
Attn(\mathbf{Q}_t, \mathbf{K}_t^u, \mathbf{V}_t^u) = \mathrm{softmax}(\frac{\mathbf{Q}_{t}^{}{\mathbf{K}_{t}^u}^\top{}}{\sqrt{d}})\mathbf{V}^u_t .
\label{equ:att1}
\end{equation}
This replacement ensures that the attention mechanism dynamically adapts to the encoded target features, thereby enhancing the precision and relevance of generated outputs.

\textbf{Denoising Target Loss.} The SAFE module aims to synthesize target images based on the information extracted from reference images. Its training objective aligns with the standard diffusion model, ensuring that the noise predicted by the model matches the added noise given the textual embeddings. Specifically, the loss function includes denoising loss and regularization loss, where the former maintains the standard objective in Eq. \ref{equ:loss} and the latter enhances the interpretability and generalizability of the embeddings. The overall training objective of the SAFE module is as follows:

\begin{equation}
\mathcal{L}^u = \mathbb{E}_{z_t^r,p^u,t,\epsilon}\big[\big|\big|\epsilon - \epsilon_\theta(z_t^r, p^u, t)\big|\big|^2\big] + ||p^u||_1 ,
\end{equation}
where $z_{t}^r$ is noise latent at timestep $t$. The former term is denoising loss and the latter term is regularization loss.

\subsection{Weakness Aware Sample Miner Module}
\label{sec:hard}
The Weakness Aware Sample Miner (WASM) module integrates downstream model feedback to synthesize crucial images, which directly address vulnerabilities of downstream models, thus enhancing its robustness and accuracy.

In stage 2 of Fig. \ref{fig:pipe}(b), the WASM module takes a real scarce image $x^r$ as input and extracts multi-layer visual features using the CLIP encoder $\Psi(\cdot)$. Unlike SAFE, which simply maps images to descriptions about targets, WASM module employs a target transfer network $\phi^c(\cdot)$ that shares the same architecture as the target extractor network but serves a distinct purpose. Through subsequent fine-tuning, $\phi^c(\cdot)$ specializes in extracting discriminative textual embeddings from target images, which explicitly encode the subtle but critical features that downstream models most frequently misclassify. 

Specifically, the target transfer network $\phi^c(\cdot)$  utilizes the pre-trained weights of the target extractor network $\phi^u(\cdot)$ to ensure training stability by preventing gradient vanishing, while retaining the semantic features learned during pre-training to prevent significant style deviations. The network then processes $\Gamma$-layer features from the CLIP visual encoder identically to SAFE, but subsequently transforms them through $\phi^c(\cdot)$ to produce crucial textual embedding $\phi^c(\Psi(x^r))$. These embeddings are finally fused them with base prompt templates to create crucial prompt $p^c$ about deceptive features.

Subsequently, the crucial prompt $p^c$ is projected into key-value pairs $\mathbf{K}_t^c$ and $\mathbf{V}_t^c$ via the learnable linear blocks $(\mathcal{W}_t^c)^K$ and $(\mathcal{W}_t^c)^V$ to compute cross-attention result activated by crucial prompt $Attn(\mathbf{Q}_t, \mathbf{K}_t^c, \mathbf{V}_t^c)$. However, images generated solely from this attention output tend to overemphasize subtle artifacts, like texture noise, causing downstream models to misclassify such features as targets, or produce images that do not contain any targets. To address this, the cross-attention output is fused with the attention result activated by $p^u$ in SAFE module to ensure semantic consistency with real targets while inserting deceptive features, as formulated below:

\begin{equation}
Attn \leftarrow \lambda \cdot Attn(\mathbf{Q}_t, \mathbf{K}_t^c, \mathbf{V}_t^c) + (1-\lambda) \cdot Attn(\mathbf{Q}_t, \mathbf{K}_t^u, \mathbf{V}_t^u),
\label{equ:attcomb}
\end{equation}
where $\lambda$ is a trade-off hyperparameter between 0 and 1. 

\textbf{Attacking Feedback Loss.} To ensure that the synthesized targets effectively challenge downstream models, the generated images $x^s$ are passed to a pre-trained downstream task model $F(\cdot)$ (e.g., a segmentation or detection model) to obtain pixel-level prediction probabilities. The goal is to maximize the difficulty distinguishing the synthesized target from the background, particularly within bounding box regions $m$. For each synthesized image $x^s$, the downstream model $F(\cdot)$ produces a pixel-level probability map $f=F(x^s)$, where $f \in R^{W \times H}$. To ensure the synthesized target blends seamlessly with background, we introduce a feedback loss designed to minimize the average prediction confidence of the target within bounding box regions $m$ relative to surrounding areas. Additionally, we also incorporate a regularization term in the training, defined as follows:

\begin{equation}
\begin{aligned}
\mathcal{L}^c 
&= ||\frac{\sum_{(i,j) \in m} f_{i,j}}{|m| + \delta }  - \frac{\sum_{(i,j) \in (1-m)} f_{i,j}}{|1-m| + \delta } ||^2 
+ ||p^c||_1 ,
\end{aligned}
\end{equation}
where $|m|$ is the number of pixels in the bounding box and $|1-m|$ is the number of pixels outside it, and $\delta $ is a small constant to avoid division by zero. Notably, we first train SAFE module, using its parameters to initialize WASM module for stable training. The WASM module is then fine-tuned to synthesise crucial targets. This two-stage method is essential because joint optimization creates competing objectives. SAFE's real target generation conflicts with WASM's undetectable generation, ultimately reducing the effectiveness of the synthetic crucial targets.

\subsection{Target Annotation Generation}
\label{sec:anno}

\begin{figure}[t!]
    \centering
    \includegraphics[width=\linewidth]{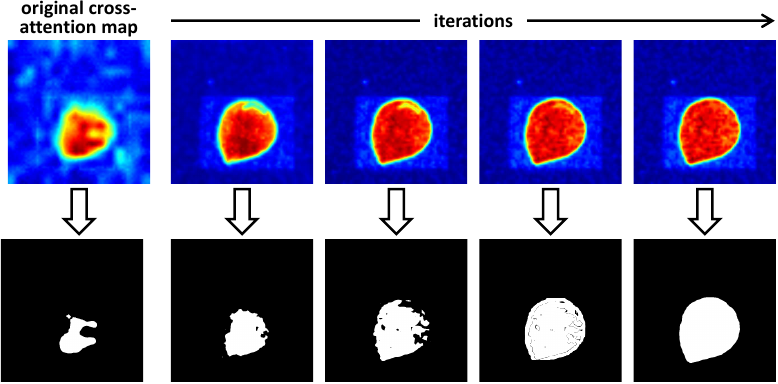} 
    \caption{Visualization of iterative refinement of the cross-attention layer. The first row displays cross-attention maps when the iteration is 0, 1, 3, 5 and 7, while the second row shows the corresponding pixel-level annotations $\hat m$.}
    
    \label{fig:anno}
\end{figure}

During inference, the direct application of cross-attention maps frequently results in insufficient fine-grained spatial details, which can compromise the accuracy of target region localization when these maps are binarized for detailed annotations, particularly in complex scenes characterized by cluttered backgrounds. By harnessing the rich spatial information encoded in self-attention maps, we can significantly enhance the spatial precision of the attention maps, thereby facilitating a more accurate delineation of target regions. Similar to \cite{sun2024iseg}, we extract self-attention maps $\mathcal{A}^{sa}$ and cross-attention maps $\mathcal{A}^{ca}$ from multiple layers of the U-Net decoder and utilize the self-attention maps to iteratively refine the cross-attention maps. In each iteration $i$, the refined cross-attention map is continuously updated as follows:

\begin{equation}
\mathcal{A}^{ca}_i = \mathcal{A}^{sa} \cdot \mathcal{A}^{ca}_{i-1}, \quad i=1,…,I,
\end{equation}
where $I$ is the total number of iterations. Notably, $\mathcal{A}^{ca}_0$ is the average of cross-attention maps activated by $p^*$ at last timestep. The normalization operation is applied after each iteration to ensure the stability of the refined map.
As shown in Fig. \ref{fig:anno}, the refinement process progressively enhances the spatial accuracy of the attention map. However, excessive iteration may lead to overfitting of local textures while ignoring global structural patterns. We therefore select $I=5$ as the optimal balance.
Finally, by applying a threshold to the refined cross-attention map, we effectively separate the target region from the background, resulting in pixel-level annotations $\hat m$ aligned with synthesized targets. 
\sethlcolor{yellow}Overall, the detailed synthesis process of images and annotations is depicted in Algo. \ref{alg:inf}.

\section{Experiment}
\subsection{Implementation Details}
Crucial-Diff is built on Stable Diffusion V1.5 \cite{rombach2022high}, training SAFE for 500 epochs and WASM for 200 epochs. The batch size is 3 and image size is 512. The optimizer AdamW utilizes a scaled learning rate initialized to 1e-6. We use a timestep of 30 and a guidance scale of 3.5 for generation. The scale factor $\sqrt{d}$ is 1 and text embedding dimension $q$ is 768. The total layer number of visual encoder $N$ is 24, the selected number $\Gamma$ is 5, $\delta$ is 0.01 and $\lambda$ is 0.5.

\begin{algorithm}[t!]
\caption{The synthesis process of proposed Crucial-Diff.}
\label{alg:inf}
\textbf{Input}: target image $x^r$, background image $x^b$, coarse mask $m$, total inference timestep $T$ \\
\textbf{Parameter}: conditional noise predictor $\epsilon_\theta(\cdot)$, VAE image encoder $\varepsilon(\cdot)$, CLIP text encoder $\tau(\cdot)$, CLIP visual encoder $\Psi(\cdot)$, target extractor network $\phi^u(\cdot)$, target transfer network $\phi^c(\cdot)$ and hyperparameter $\lambda$ \\
\textbf{Output}: the synthetic image $x^s$, corresponding annotation $\hat{m}$\\

\begin{algorithmic}[1] 
\STATE $z_T^s \sim \mathcal{N}(0,I)$; \quad $z_T^s = \textbf{concat}\big(z_T^s,m,\varepsilon\big(x^b(1-m)\big)\big)$
\STATE Select $\Gamma$ textual embeddings from $\phi^u(\Psi(x^r))$.
\STATE $p^u = \left[\tau(\text{``a photo of''}); \phi^u(\Psi_1(x^r)); \ldots; \phi^u(\Psi_\Gamma(x^r))\right]$
\STATE $p^c = \left[\tau(\text{``a photo of''}); \phi^c(\Psi_1(x^r)); \ldots; \phi^c(\Psi_\Gamma(x^r))\right]$
\FOR{$t = T$ to $1$}
    \FOR{each cross attention layer}
        \STATE Compute $\mathbf{K}_t^u$, $\mathbf{V}_t^u$ using SAFE module.
        \vspace{2pt}
        \STATE $\mathbf{K}_t^u = (\mathcal{W}_t^u)^K \cdot p^u$; \quad $\mathbf{V}_t^u = (\mathcal{W}_t^u)^V \cdot p^u$  \\
        \vspace{2pt}
        \STATE $Attn(\mathbf{Q}_t, \mathbf{K}_t^u, \mathbf{V}_t^u) = \mathrm{softmax}(\frac{\mathbf{Q}_{t}^{}{\mathbf{K}_{t}^u}^\top{}}{\sqrt{d}})\cdot\mathbf{V}^u_t$ \\
        \vspace{2pt}
        \STATE Compute $\mathbf{K}_t^c$, $\mathbf{V}_t^c$ using WASM module.
        \vspace{2pt}
        \STATE $\mathbf{K}_t^c = (\mathcal{W}_t^c)^K \cdot p^c$; \quad $\mathbf{V}_t^c = (\mathcal{W}_t^c)^V \cdot p^c$ 
        \vspace{2pt}
        \STATE $Attn(\mathbf{Q}_t, \mathbf{K}_t^c, \mathbf{V}_t^c) = \mathrm{softmax}(\frac{\mathbf{Q}_{t}^{}{\mathbf{K}_{t}^c}^\top{}}{\sqrt{d}})\cdot\mathbf{V}^c_t$ 
        \vspace{2pt}
        \STATE Update $Attn \leftarrow \lambda \cdot Attn(\mathbf{Q}_t, \mathbf{K}_t^c, \mathbf{V}_t^c) + (1-\lambda) \cdot Attn(\mathbf{Q}_t, \mathbf{K}_t^u, \mathbf{V}_t^u)$ 
        \vspace{2pt}
    \ENDFOR
    \STATE Compute $z_{t-1}^s$ according to \textbf{Eq. \ref{equ:reverse}}.
    \IF{$t=1$}
        \STATE Compute self-attention maps $\mathcal{A}^{sa}$
        \vspace{2pt}
        \STATE $\mathcal{A}^{ca} = \lambda\mathrm{softmax}(\frac{\mathbf{Q}_{t}^{}{\mathbf{K}_{t}^c}^\top{}}{\sqrt{d}}) + (1 - \lambda)\mathrm{softmax}(\frac{\mathbf{Q}_{t}^{}{\mathbf{K}_{t}^u}^\top{}}{\sqrt{d}})$
        \FOR{$i = I$ to $1$} 
            \STATE $\mathcal{A}^{ca} = \mathcal{A}^{sa} \cdot \mathcal{A}^{ca}$
        \ENDFOR
       \STATE $\hat{m} = \text{Binarize}(\mathcal{A}^{ca})$ 
    \ENDIF
\ENDFOR
\STATE Use VAE image decoder to recover $z_0$ into the image $x^s$.
\STATE \textbf{return} $x^s$, $\hat{m}$
\end{algorithmic}
\end{algorithm}

\subsection{Datasets and Metrics}
The experiments assess training efficiency, generation quality and downstream task performance in two real-world scenarios, industrial inspection and medical diagnosis.

\textbf{Industrial inspection.} We evaluate Crucial-Diff on the MVTec dataset \cite{bergmann2019mvtec}, which contains 3,629 normal training images and a test set of 467 normal and 1,258 anomaly images with mask labels. Additionally, we evaluate our method on the Real-IAD dataset \cite{wang2024real}. This dataset contains 150K images covering 30 different object categories, providing more diverse and challenging real-world industrial scenarios. Following \cite{hu2023anomalydiffusion}, one-third of the anomaly images are used for training, while the rest is used to test downstream tasks. Generation performance is evaluated with Intra-cluster pairwise LPIPS distance (IC-LPIPS) \cite{ojha2021few}, Inception Score (IS) \cite{salimans2016improved} and Kernel Inception Distance (KID) \cite{binkowski2018demystifying}. Detection performance is evaluated with Average Precision (AP) and maximum F1 score (F1-MAX). Computational efficiency is measured by metric Time, which means total training hours for all targets.

\textbf{Medical Diagnosis.} We conduct experiments on five publicly available polyp segmentation datasets: ETIS \cite{silva2014toward}, CVC-ClinicDB/CVC-612 \cite{bernal2015wm}, CVC-ColonDB \cite{tajbakhsh2015automated}, EndoScene \cite{vazquez2017benchmark} and Kvasir \cite{jha2020kvasir}. The training set includes Kvasir with 900 samples and CVC-ClinicDB with 550 samples. while the test set comprises CVC-300, CVC-ClinicDB, CVC-ColonDB, ETIS and Kvasir with a total of 898 samples.
The evaluation uses IC-LPIPS, IS and KID to assess synthesized image diversity and quality, along with mIoU and mDice for downstream accuracy. Higher values denote better performance for all metrics except KID, where lower values are optimal.

\begin{figure*}[t!]
    \centering
    \includegraphics[width=\linewidth]{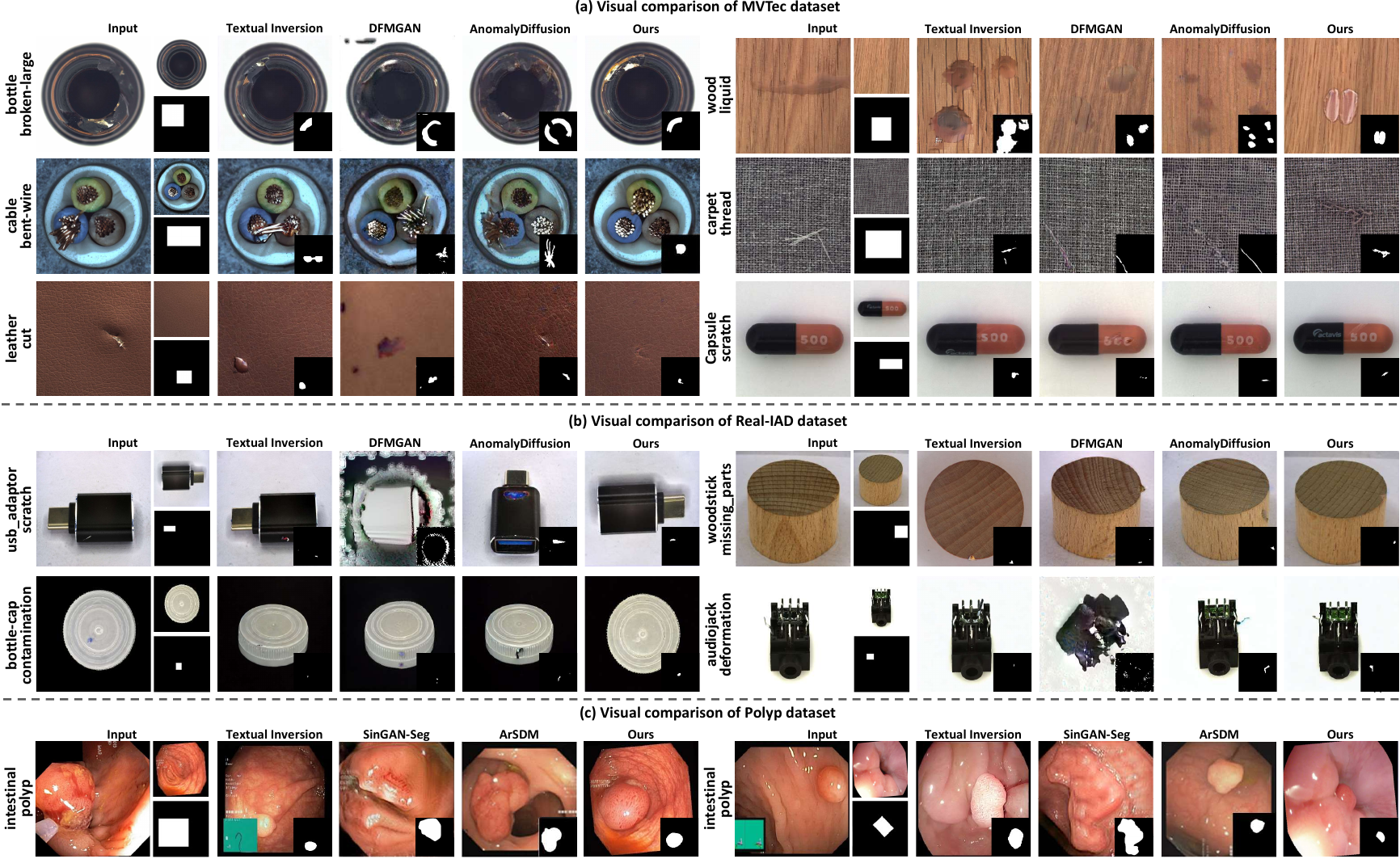} 
    \caption{Visualization comparison of synthetic images generated by Crucial-Diff and other methods. (a) is MVTec result, (b) is Real-IAD dataset result and (c) is polyp dataset result with the annotations of synthetic targets at bottom right of each image. Our generated samples exhibit features like seamless background fusion, surface irregularities or subtle color variations that effectively disrupt downstream models and significantly increase the difficulty of target distinction.}
    \label{fig:visual}
\end{figure*}

\subsection{Qualitative Comparison} 
\subsubsection{Comparison of Synthetic Anomalies} 
As shown in Fig. \ref{fig:visual}(a), in MVTec dataset, previous methods, such as AnomalyDiffusion \cite{hu2023anomalydiffusion}, tend to generate targets that are visually distinct and easily recognizable, limiting their effectiveness in improvement of downstream models. In contrast, Crucial-Diff excels at generating subtle anomalies, such as faint cuts and barely visible breaks. These subtle targets are easily overlooked by downstream tasks as training data can effectively improve detection results. This advantage is further validated on the Real-IAD dataset, as illustrated in Fig. \ref{fig:visual}(b). While other methods, particularly DFMGAN \cite{duan2023few}, struggle to restore the original appearance of objects, our method successfully generates realistic and subtle damage.

In Fig. \ref{fig:visual}(c), on the polyp dataset, Crucial-Diff synthesizes polyps with blurry boundaries and low contrast, allowing them to blend naturally into the surrounding tissue. This not only improves visual realism but also enhances the model’s applicability in clinical scenarios. In comparison, previous methods often produce polyps with sharp edges and high contrast. While such synthetic samples are easier to detect, they lack anatomical plausibility and reduce practical utility.

\subsubsection{Distribution Analysis of Synthetic Anomalies} 
\begin{figure}[t!]
    \centering
    \setlength{\abovecaptionskip}{-0cm}
    \setlength{\belowcaptionskip}{-0.77cm} \includegraphics[width=0.92\linewidth]{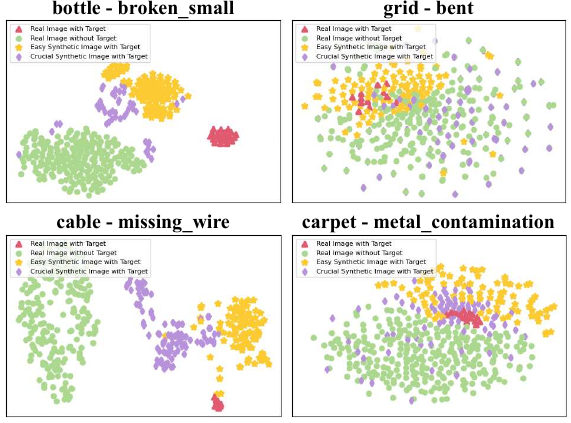} 
    \captionsetup{font={scriptsize}}
    \caption{T-SNE visualization of different targets. The pink triangle represents a real image with target, the green circle represents a real image without target, the yellow pentagram represents an easy synthetic image with target, and the purple diamond represents a crucial synthetic image with target.}
    \label{fig:tsne}
\end{figure}

The t-SNE visualization in Fig. \ref{fig:tsne} further verifies the distribution of crucial samples in Fig. \ref{fig:intro}(a). The visualization demonstrates that crucial samples exhibit closer alignment with the distribution of real non-target images, whereas simple samples display substantial overlap with real target image distributions. This fundamental divergence explains the inherent limitations of previous synthesis methods in improving downstream model performance. Significantly, the results demonstrated that Crucial-Diff has a unique ability to synthesise crucial samples for downstream detection models.

\subsubsection{Visualization of Segmentation Results}
\begin{figure}[t!]
    \centering
    \setlength{\abovecaptionskip}{-0cm}
    \setlength{\belowcaptionskip}{-0.77cm} \includegraphics[width=0.99\linewidth]{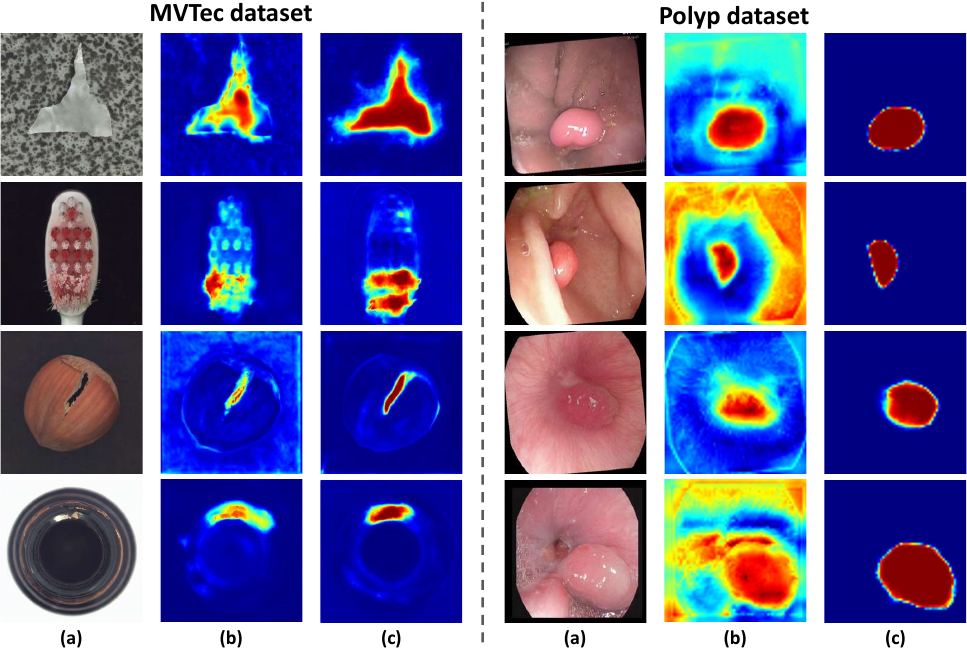} 
    \captionsetup{font={scriptsize}}
    \caption{Visualization comparison of segmentation results on different training data. (a) shows samples generated by our method. (b) shows the segmentation results of a downstream model trained on one-third of the original dataset, while (c) shows the segmentation results of a downstream model augmented with our 1000 key samples.}
    \label{fig:seg}
\end{figure}

To further explore the characteristics of crucial samples, we provide a visualization of segmentation results. Fig. \ref{fig:seg}(a) displays a representative critical sample synthesized by our method, (b) presents the segmentation output of a downstream model trained on only one-third of the original dataset and (c) shows the result from a downstream model trained on 1000 critical samples generated by our approach, evaluated on the same sample.
The results demonstrate that training with critical samples enables the model to retain high segmentation sensitivity, particularly toward subtle anomalous features that are often overlooked. By incorporating challenging features designed to target under-learned regions of the decision boundary, critical samples effectively expand the coverage of hard examples beyond what conventional training data provide. Consequently, this approach leads to substantial improvements in the model’s generalization capacity and robustness when deployed in complex and variable environments.

\begin{table*}[t!]
    \centering
    \belowrulesep=0pt
    \aboverulesep=0pt
    \setlength{\tabcolsep}{13.5pt}
    \caption{Comparison with other generation methods in MVTec and Real-IAD, where \textit{Trad.} denotes traditional augmentation methods, \textit{Gen.} denotes generative model based methods, \textit{Uni.} denotes unified methods and \textit{Ano.Diff.} denotes AnomalyDiffusion \cite{hu2023anomalydiffusion}. Metric Time refers to the total training hour on a single dataset, while $\dagger$ represents the training time across all datasets.}
    \begin{tabular}{c|c|cccc|cccc}
    \toprule[1.5pt]
        & \multirow{2}{*}{\textbf{Methods}}  &  \multicolumn{4}{c|}{\textbf{MVTec}}  & \multicolumn{4}{c}{\textbf{Real-IAD}} \\
        \cmidrule(lr){3-6} \cmidrule(l){7-10}
        &   &Time$\downarrow$ & IS$\uparrow$ & IC-L$\uparrow$ & KID$\downarrow$ &Time$\downarrow$ & IS$\uparrow$ & IC-L$\uparrow$ & KID$\downarrow$ \\
    \toprule[1.5pt]
         \multirow{2}{*}{\textit{Trad.}} &  Rotate  &- 
& \textbf{1.87} & 0.16 & - &- &  \textbf{2.24} & 0.08  & - \\
         &  Crop-Paste\cite{lin2021few}  &- &  1.51&  0.14& - & - & 2.05 & 0.04  & -  \\
    \hline
         \multirow{3}{*}{\textit{Gen.}} &  Textual Inversion\cite{gal2023image} & 113.49 &  1.66 & 0.24 & \textbf{0.07} & 224.62 &  1.31 & 0.03 & 0.11 \\
         &  DFMGAN\cite{duan2023few}  &289.50 &  1.72&  0.20&  0.08  & 518.82 & 1.79 & \textbf{0.13} & 0.24  \\
         &  \textit{Ano.Diff.}\cite{hu2023anomalydiffusion}  &172.43 &  1.80 & 0.32 &  0.13 & 206.98 & 	1.90 & 0.04 & \textbf{0.10}\\

    \hline
         \textit{Uni.} & Ours  &\textbf{61.15}$^{\dagger}$  & 1.84 & \textbf{0.33} & \textbf{0.07} & \textbf{61.15}$^{\dagger}$ & 1.92 & 0.05 & \textbf{0.10} \\
    \toprule[1.5pt]
    \end{tabular}
    
    \label{tab:genm}
\end{table*}

\subsection{Quantitative Comparison}
\subsubsection{Comparison with Generation Methods}
As summarized in Table \ref{tab:genm}, our proposed method achieves an IS of 1.84 on the MVTec dataset, surpassing AnomalyDiffusion \cite{hu2023anomalydiffusion} at 1.80 and Textual Inversion \cite{gal2023image} at 1.66. Although the rotational baseline achieves the highest IS due to its close distributional alignment with the real training data, our method delivers more comprehensive performance across diverse evaluation metrics. Furthermore, Crucial-Diff reduces the KID to 0.07 and elevates the IC-LPIPS to 0.33, validating that Crucial-Diff can effectively optimise diversity and fidelity simultaneously. While DFMGAN achieves a high IC-LPIPS on Real-IAD by generating distorted and unrealistic images, our method maintains a better balance between diversity and semantic authenticity.

\begin{table}[t!]
    \centering
    \belowrulesep=0pt
    \aboverulesep=0pt
    \setlength{\tabcolsep}{6.5pt}
    \caption{Comparison with other generation methods in five polyp datasets (CLINICDB, ETIS, ENDOSCENE, KVASIR, COLONDB). \textit{Trad.} denotes traditional augmentation methods, \textit{Gen.} denotes generative model based methods, \textit{Uni.} denotes unified methods. Metric Time refers to the total training hour, while $\dagger$ represents the training time across all datasets.}
    \begin{tabular}{c|c|cccc}
    \toprule[1.5pt]
         &  Methods &Time$\downarrow$
& IS$\uparrow$ & IC-L$\uparrow$ & KID$\downarrow$ \\
    \toprule[1.5pt]
         \multirow{2}{*}{\textit{Trad.}} &  Rotate &- 
& \textbf{3.74} & 0.27 & -  \\
         &  Crop-Paste\cite{lin2021few} &- 
& 3.52 & 0.27  & -  \\
    \hline
         \multirow{2}{*}{\textit{Gen.}} &  Textual Inversion\cite{gal2023image} & \textbf{16.88} & 3.54 & 0.13 & 0.06 \\
         & SinGAN-Seg\cite{thambawita2022singan}  &873.75 
&  3.55 &  0.54 & 0.07  \\
         &  ArSDM\cite{du2023arsdm} &4818.33 
&  3.00 & 0.56 & 0.09  \\
    \hline
         \textit{Uni.} & Ours  & 61.15$^{\dagger}$ & 3.59 & \textbf{0.59} & \textbf{0.05} \\
    \toprule[1.5pt]
    \end{tabular}
    
    \label{tab:genp}
\end{table}

\begin{table}[t!]
    \setlength{\tabcolsep}{16pt}
    \centering
    \belowrulesep=0pt
    \aboverulesep=0pt
     \caption{Comparison of mask annotation accuracy across Methods. Quantitative evaluation of synthetic mask quality is performed by computing the IoU between expert-annotated ground truth masks and those generated by different methods. Higher IoU values indicate superior alignment and annotation accuracy.} 
    \begin{tabular}{c|ccc}
        \toprule[1.5pt]
        Methods & \textbf{MVTec} & \textbf{Real-IAD} & \textbf{Polyp} \\
        \hline
        \textit{Ano.Diff.} & 36.73\% & 30.99\% & -\\
        ArSDM & - & - & 83.32\% \\
        \hline
        Ours & \textbf{63.48\%} & \textbf{44.35\%} & \textbf{86.71\%} \\
        \toprule[1.5pt]
    \end{tabular}
   
    \label{tab:acc}
\end{table}

Crucial-Diff also exhibits consistently strong performance on the polyp dataset, as displayed in Table \ref{tab:genp}. Our method achieves an IS of 3.59, a KID of 0.05 and an IC-LPIPS score of 0.59. Notably, the minimal training time of Textual Inversion \cite{gal2023image} can be attributed to the fact that it only learns a single target feature, while our method adopts comprehensive training covering all three datasets.  These results show that Crucial-Diff generalizes well across medical domains, generating crucial images that effectively improve downstream model performance while significantly reducing training time.

Furthermore, to validate the accuracy of synthetic targets and their segmentation masks, we conduct a systematic evaluation across three public datasets, where ten generated images per category are randomly selected and annotated in detail by domain experts. The IoU is used as the evaluation metric to measure the accuracy of the generated results by calculating the similarity between the generated mask and the expert-drawn mask. 
As summarized in Table \ref{tab:acc}, our approach consistently outperforms state-of-the-art methods such as AnomalyDiffusion \cite{hu2023anomalydiffusion} and ArSDM \cite{du2023arsdm} across all datasets, achieving higher IoU scores. These results confirm the high annotation reliability of our generated samples, demonstrating its suitability for training robust and generalizable downstream models.

\subsubsection{Comparison of Synthetic Data Effectiveness}
In industrial anomaly detection, this study systematically evaluates the proposed synthetic data generation method using both U-Net and PRN \cite{zhang2023prototypical} architectures on the MVTec dataset. The Real-IAD dataset is excluded from the comparative study since most generative methods exhibit limited generation quality on this benchmark, making downstream performance comparisons less meaningful. As summarized in Table \ref{tab:addm}, the proposed approach consistently enhances performance across both models. On MVTec, U-Net trained with the synthetic samples achieves 83.6\% AP and 78.1\% F1-MAX, surpassing all competing methods. The PRN \cite{zhang2023prototypical} method also shows clear gains when augmented with the synthetic data, demonstrating robust cross-architectural generalization.

For medical image segmentation, all experiments are conducted on a polyp dataset comprising five public datasets. As shown in Table \ref{tab:addp}, SegFormer trained with our synthetic data achieves 87.7\% mDice and 81.6\% mIoU, while CTNet \cite{xiao2024ctnet} augmented with our data reaches 88.3\% mDice and 82.1\% mIoU. Both models outperform those trained on data generated by previous methods, further demonstrating the effectiveness of our synthetic data generation approach.

\begin{table}[t!]
    \setlength{\tabcolsep}{19pt}
    \centering
    \belowrulesep=0pt
    \aboverulesep=0pt
    \caption{Performance of different downstream models trained on synthetic datasets generated by various methods. Each generation method produced 1,000 images on MVTec and Real-IAD datasets. Evaluation using pixel-level AP and F1-MAX metrics. The highest score is highlighted in \textbf{bold}.}
    \begin{tabular}{lcc}
    \toprule[1.5pt]
    
    Methods & AP$\uparrow$  & F1-MAX$\uparrow$ \\
    \toprule[1.5pt]
    U-Net \\
        + rotate & 57.6 & 57.0  \\
        + Crop-Paste\cite{lin2021few} & 74.1 & 71.9  \\
        + Textual Inversion\cite{gal2023image} & 71.1 & 66.5  \\
        + DFMGAN\cite{duan2023few} & 63.6 & 64.5\\
        + \textit{Ano.Diff.}\cite{hu2023anomalydiffusion}  & 82.2 & 77.1 \\
        + Ours  & \textbf{83.6} & \textbf{78.1}   \\
    \hline
    PRN\cite{zhang2023prototypical} \\
        + rotate & 47.3 & 50.3  \\
        + Crop-Paste\cite{lin2021few} & 50.4 & 51.8   \\
        + Textual Inversion\cite{gal2023image} & 46.2 & 48.9 \\
        + DFMGAN\cite{duan2023few} & 39.3 & 43.2 \\
        + \textit{Ano.Diff.}\cite{hu2023anomalydiffusion}  & 58.6   & 53.2 \\
        + Ours  & \textbf{63.7} & \textbf{62.0}  \\
    \toprule[1.5pt]
    \end{tabular}
    
    \label{tab:addm}
\end{table}

\begin{table}[t!]
    \centering
    \belowrulesep=0pt
    \aboverulesep=0pt
    \setlength{\tabcolsep}{19pt}
    \caption{Performance of different downstream models trained on synthetic datasets generated by various methods. Each generation method produced 1,450 images on five datasets (CLINICDB, ETIS, ENDOSCENE, KVASIR, COLONDB). Evaluation using average mDICE and mIOU metrics across five datasets. The highest score is highlighted in \textbf{bold}.}
    \begin{tabular}{lcc}
    \toprule[1.5pt]
    \textbf{Methods} & \textbf{mDice$\uparrow$ }  & \textbf{mIoU$\uparrow$ }\\ 
    \toprule[1.5pt]
    Segformer &   \\
        + rotate  & 81.6 & 74.0 \\
        + Crop-Paste\cite{lin2021few}  & 80.9 & 73.2\\
        + Textual Inversion\cite{gal2023image}  & 67.8 & 59.2\\
        + SinGAN-Seg\cite{thambawita2022singan} & 79.5 & 71.7 \\        
        + ArSDM\cite{du2023arsdm}  & 82.4 & 75.2 \\
        + Ours  &  \textbf{87.7} & \textbf{81.6} \\

    \hline
    CTNet\cite{xiao2024ctnet} &  \\
        + rotation   & 87.0 & 80.4\\
        + Crop-Paste\cite{lin2021few}  &  86.7 &  80.2 \\
        + Textual Inversion\cite{gal2023image}  &  68.5 & 63.6 \\
        + SinGAN-Seg\cite{thambawita2022singan} & 68.4 &	78.5 \\
        + ArSDM\cite{du2023arsdm}   & 87.2 &  80.7 \\
        + Ours  & \textbf{88.3} & \textbf{82.1} \\

    \toprule[1.5pt]
    \end{tabular}
    
    \label{tab:addp}
\end{table}

\subsubsection{Comparison with Downstream Task Methods}
To evaluate the effectiveness of our Crucial-Diff framework in improving downstream task performance, we compare a U-Net model trained on data generated by our method against state-of-the-art methods. In Table \ref{tab:downm}, our Crucial-Diff achieves pixel-level AP of 84.2\% and F1-MAX of 78.4\%, outperforming both unsupervised and supervised SOTA methods. Notably, compared to PRN \cite{zhang2023prototypical}, which has AP of 78.6\% and F1-MAX of 62.3\%, our approach significantly improves detecting subtle anomalies.
As for the polyp dataset, our Crucial-Diff achieves mDice of 87.9\% and mIoU of 81.9\%, surpassing the previous method CTNet \cite{fan2020pranet} by 1\% in mIoU and 1.2\% in mDice, as shown in Table \ref{tab:downp}. Empirical results demonstrate Crucial-Diff's consistent performance gains across industrial and medical domains, confirming its task-agnostic generalization capability for diverse downstream applications.

\begin{table}[t!]
\setlength{\tabcolsep}{10pt}
\centering
\belowrulesep=0pt
\aboverulesep=0pt
\caption{Comparison of best anomaly detection performance (pixel-level / image-level) between previous methods and a U-Net model trained on data generated by our method on MVTec dataset. Methods marked with * indicate our reproduced and marked in \textbf{bold} and \underline{underline} indicate the highest and second-highest scores.}
\begin{tabular}{c|c|cc}
\toprule[1.5pt]
    & Methods  & AP$\uparrow$ & F1-MAX$\uparrow$  \\
\toprule[1.5pt]
\multirow{3}{*}{Unsupervised} 
     &  RD4AD\cite{Deng_2022_CVPR} & 48.6 / 96.5 & 53.8 / 95.2 \\
     & SimpleNet\cite{liu2023simplenet} &  45.9 / 98.4 & 49.7 / 95.8 \\
     & ViTAD\cite{zhang2023exploring} & 55.3 / \underline{99.4}  & 58.7 / \underline{97.3}   \\
\hline
\multirow{4}{*}{Supervised} 
     &  DevNet*\cite{pang2021explainable} & 49.3 / 98.6  & 42.8 / 97.2  \\
     &  DRA*\cite{ding2022catching} &  25.7 / 95.8 & 43.4 / 94.7  \\
     &  PRN*\cite{zhang2023prototypical} &  \underline{78.6} / 96.8 &  \underline{62.3} / 93.4  \\ 
     
     &  Ours& \textbf{84.2} / \textbf{99.8} & \textbf{78.4} / \textbf{98.9}  \\

\toprule[1.5pt]
\end{tabular}

\label{tab:downm}
\end{table}

\begin{table}[t]
    \centering
    \setlength{\tabcolsep}{6pt}
    \belowrulesep=0pt
    \aboverulesep=0pt
    \caption{Segmentation comparisons (mDice / mIoU) between previous methods and a SegFormer \cite{xie2021segformer} model trained on data generated by our method on polyp datasets. Methods marked in \textbf{bold} and \underline{underline} indicate the highest and second-highest scores.}
    \begin{tabular}{c|cccc}
    \toprule[1.5pt]

        ~ & PraNet\cite{fan2020pranet} & PVT\cite{dong2021polyp} & CTNet\cite{xiao2024ctnet} & Ours \\ 
        \hline
        ClinicDB & 89.9 / 84.9 & \textbf{93.7} / \textbf{88.9} & 93.0 / 88.4 & \underline{93.5} / \underline{88.7}\\ 
        
        ETIS & 62.8 / 56.7 & 78.7 / 70.6 & \underline{78.9} / \underline{71.4} & \textbf{81.7} / \textbf{74.3}\\ 
        
        EndoScene & 87.1 / 79.7 & \underline{90.0} / 83.3 & \underline{90.0} / \underline{83.6} & \textbf{91.8} / \textbf{86.7} \\ 
        
        Kvasir & 89.8 / 84.0 & \underline{91.7} / \textbf{86.4} & 91.3 / \underline{86.3} & \textbf{92.1} / \textbf{86.4}\\ 
        
        ColonDB & 70.9 / 64.0 & \underline{80.8} / 72.7 & \textbf{81.2} / \textbf{73.8} & 80.4 / \underline{73.6}\\ 
        \hline
        Average & 80.1 / 73.9 & 87.0 / 80.4 & 86.9 / 80.7 & \textbf{87.9} / \textbf{81.9}\\ 
    \toprule[1.5pt]
    \end{tabular}

\label{tab:downp}
\end{table}

\subsection{Ablation Study}
\subsubsection{Effect of Different Components}
To assess the contributions of each module in Crucial-Diff, we compare three variants: 
1) SAFE Only, which uses only the SAFE module for generating samples; 2) WASM Only, which relies exclusively on the WASM module without initialization from the SAFE module; and 3) SAFE + WASM (Ours), combining both modules to generate crucial samples.
Table \ref{tab:component} demonstrates the impact of image training on downstream tasks generated under different settings. Only with SAFE module, AP drops by 5.80\% and F1-MAX by 4.62\%, which means that using only simple data does not allow the downstream model to detect the target better. Only with SRAM module reduces AP by 5.06\% and F1-MAX by 1.49\%, as it fails to generate target images, forcing the downstream model to learn from non-target samples and misidentify non-target regions. In contrast, our method achieves the highest AP and F1-MAX, demonstrating the necessity of each module.

\begin{table}[t!]
    \setlength{\tabcolsep}{7pt}
    \centering
    \caption{Ablation study on main components contributions in Crucial-Diff. The MVTec uses pixel-level AP and F1-MAX. The polyp dataset uses the average results of mIoU and mDice.}
    \begin{tabular}{ccccccc}
    \toprule[1.5pt]
        ~ & \multicolumn{2}{c}{Module} & \multicolumn{2}{c}{MVTec} &  \multicolumn{2}{c}{Polyp} \\ 
        \cmidrule(lr){2-3} \cmidrule(lr){4-5} \cmidrule(lr){6-7}
        ~ & SAFE & WASM & AP$\uparrow$ & F1-MAX$\uparrow$  & mDice$\uparrow$ & mIoU$\uparrow$ \\
        \hline 
        1) & \ding{52} & ~ & 77.83 & 73.50 & 71.90 &	59.65 \\
        2) & & \ding{52} & 78.57 & 76.63 &  73.41 & 66.29 \\ 
        3) & \ding{52} & \ding{52} & \textbf{83.63} & \textbf{78.12} & \textbf{87.69} & \textbf{81.64} \\
    \toprule[1.5pt] 
    \end{tabular}
    
    \label{tab:component}
\end{table}

\begin{table}[t!]
    \setlength{\tabcolsep}{17.5pt}
    \centering
     \caption{Ablation study of the generated image quality on MVTec for different number of selected text embeddings value $\Gamma$.}
    \begin{tabular}{c|cccc}
    \toprule[1.5pt]
        ~ & $\Gamma$=1 & $\Gamma$=3 & $\Gamma$=5 & $\Gamma$=7 \\
        \hline
        IS & 1.77 & 1.79 & \textbf{1.84} & \textbf{1.84} \\
        KID & 0.11 & 0.11 & \textbf{0.07} & 0.08 \\

    \toprule[1.5pt]
    \end{tabular}
   
    \label{tab:k_ablation}
\end{table}

\begin{table}[t!]
    \setlength{\tabcolsep}{21pt}
    \centering
     \caption{Ablation of fusion strategies on MVTec. Comparative analysis of three paradigms: 1) feature level addition, 2) KV concatenation, and 3) our output integration approach.} 
    \begin{tabular}{c|ccc}
    \toprule[1.5pt]
        Methods & IS$\uparrow$ & IC-L$\uparrow$ & KID$\downarrow$ \\
        \hline
        1) & 1.57 & 0.23 & 0.07\\
        2) & 1.59 & 0.24 & \textbf{0.06}\\
        \hline 
        3) & \textbf{1.84} & \textbf{0.33} & 0.07\\
    \toprule[1.5pt]
    \end{tabular}
   
    \label{tab:attn_ablation}
\end{table}

\subsubsection{Number of Target Text Embeddings}
To investigate the impact of the number of hierarchical features extracted from the CLIP image encoder, we conduct an ablation study on the number of target text embeddings $\Gamma$. The quantitative results are presented in Table \ref{tab:k_ablation}. As \( \Gamma \) increases, the generation performance improves until reaching an optimal configuration at 5, where the synthetic images achieve the superior IS and KID. Beyond $\Gamma = 5$, the improvements become marginal, leading us to select 5 as the optimal number of layers. This configuration effectively balances low-level details and high-level semantics, facilitating the synthesis of diverse and realistic images.

\subsubsection{Fusion Strategies of cross attention}
To validate the design of the fusion operation within the cross-attention mechanism, we conduct a comparative ablation: 1) feature level addition: $\mathbf{V}_t \leftarrow \lambda\cdot\mathbf{V}^c_t + (1-\lambda)\cdot\mathbf{V}^u_t$ and $\mathbf{K}_t \leftarrow \lambda\cdot\mathbf{K}^c_t + (1-\lambda)\cdot\mathbf{K}^u_t$; 2) KV concatenation: $Attn \leftarrow Attn(\mathbf{Q}_t, [\mathbf{K}_t^c;\mathbf{K}_t^u], [\mathbf{V}_t^c;\mathbf{V}_t^u])$; and 3) our output integration approach. Table \ref{tab:attn_ablation} demonstrate the superior performance of our approach. Our method improves IS by 0.27 over feature addition and 0.25 over KV concatenation. It also increases IC-LPIPS by 0.10 and 0.09 over the two baselines, respectively. The strategy we propose enables stable and controllable integration of deceptive features and semantic constraints, whereas other fusion methods lead to training instability issues.

\subsubsection{Value of Trade-off Hyperparameter $\lambda$}
Our ablation study on the trade-off parameter $\lambda$ demonstrates the robustness of the proposed method across different weightings (0.1-0.9) on the MVTec dataset. As shown in Table \ref{tab:lambda_ablation}, both IS and KID remain stable, indicating minimal sensitivity to parameter changes. The consistent KID scores particularly highlight our network's adaptive capability to maintain distribution alignment regardless of $\lambda$ values. This is because $\lambda$ is integrated into the network training process rather than directly controlling the output results. This design ensures stability in the early stages of training while making the model insensitive to specific values of $\lambda$, thereby making the model more robust.

\begin{table}[t!]
    \setlength{\tabcolsep}{11.5pt}
    \centering
     \caption{Ablation study on the generated image quality using MVTec dataset for different values of trade-off hyperparameter $\lambda$.}
    \begin{tabular}{c|ccccc}
    \toprule[1.5pt]
        ~ & $\lambda$=0.1 & $\lambda$=0.3 & $\lambda$=0.5 & $\lambda$=0.7  & $\lambda$=0.9 \\
        \hline
        IS & 1.83 & 1.84 & 1.84 & 1.85 & 1.85 \\
        KID & 0.07 & 0.07 & 0.07 & 0.08 & 0.09 \\
    \toprule[1.5pt]
    \end{tabular}
    \label{tab:lambda_ablation}
\end{table}

\begin{figure}[t!]
    \centering
    \includegraphics[width=\linewidth]{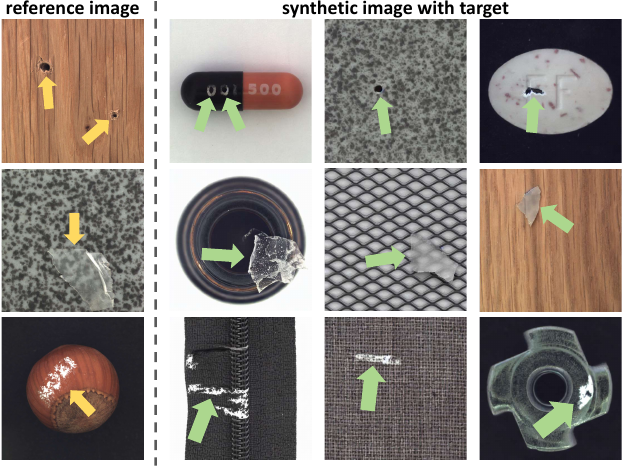} 
    \caption{Visualization of cross-category transferability of our Crucial-Diff, which generates targets not present in the original dataset. Yellow arrows indicate targets in reference images, while green arrows highlight targets generated by our method.}
    \label{fig:trans}
\end{figure}

\section{Discussion}
\subsection{Cross-Category Transferability}
In our experiments, we found that our Crucial-Diff demonstrates significant cross-category transferability, allowing the synthesis of one target into a completely different scene. For instance, as shown in Fig. \ref{fig:trans}, we successfully synthesized a ``hole" defect, typically associated with wood, into a ``capsule". This capability arises from two design choices: encoding visual features into a domain-agnostic textual embedding space to decouple target characteristics from their context and training a unified model on data from multiple domains to establish semantic associations across different scenes. As a result, we can generate targets on objects that have never been observed before, addressing dataset gaps. This capability is crucial for practical applications, especially in data-limited scenarios, as it enhances synthetic data diversity and boosts downstream model robustness.

\begin{figure}[t!]
    \centering
    \setlength{\abovecaptionskip}{-0cm}
    \setlength{\belowcaptionskip}{-0.45cm} 
    \includegraphics[width=0.98\linewidth]{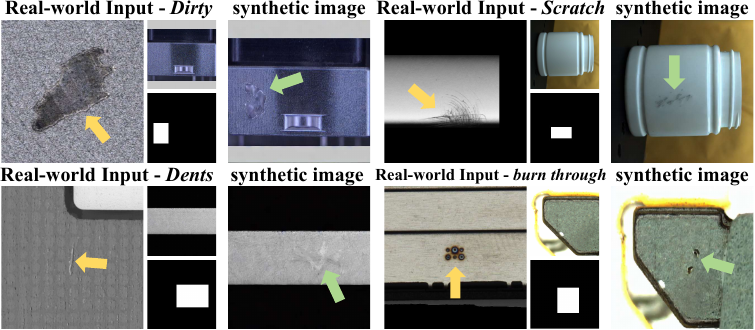} 
    \captionsetup{font={scriptsize}}
    \caption{Synthesis results using real-world images as references. The left column shows the target samples never previously observed in the destination background, while the right column presents our synthesized outputs.} 

    \label{fig:real}
\end{figure}

\begin{figure}[t!]
    \centering
    \includegraphics[width=\linewidth]{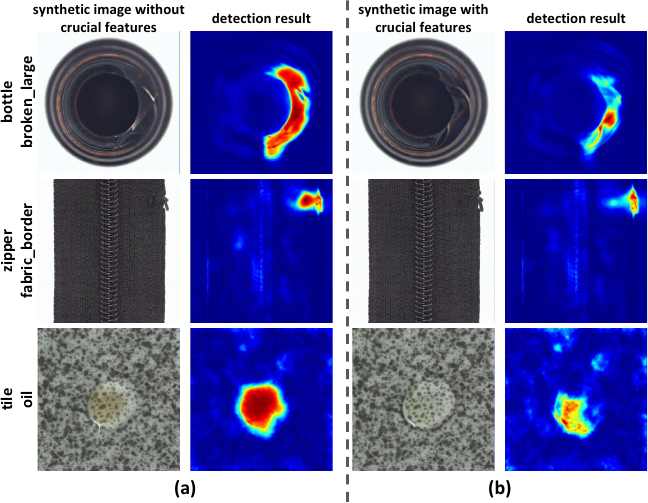} 
    \caption{Illustration of special crucial images. (a) Easy synthetic images without crucial features are effortlessly detected by the downstream model, demonstrating high prediction scores. (b) Special crucial synthetic images, despite their visual similarity to easy samples, are challenging for the downstream model to detect.}
    
    \label{fig:crucial_feature}
\end{figure}

\subsection{Real-world Images Generation Results}
To systematically evaluate model generalization, we collect out-of-distribution (OOD) defect samples from real-world industrial scenarios as reference inputs. As illustrated in Fig. \ref{fig:real}, our synthesis framework achieves: 1) precise reconstruction of unseen defect types, and 2) photorealistic rendering across domain-shifted backgrounds. Quantitative experiments confirm our framework synthesizes high-fidelity targets (e.g., dents) unseen in training, relying solely on reference images. The framework further generalizes to novel target-background combinations, generating  a wide range of valuable training samples for downstream models.

\subsection{Special Cases of Synthetic Crucial Images}
Most of the generated crucial images are significantly different in appearance from easy synthetic images. However, interestingly, we observe that some crucial images appear almost identical to simple images in appearance, yet the prediction probabilities of the downstream task decline significantly.
As shown in Fig. \ref{fig:crucial_feature}, two nearly identical broken bottle images yield drastically different prediction probabilities in the specific downstream model. 
This extends our assumption that crucial images not only exhibit subtle defects but generate patterns highly similar to easily detectable images. However, these crucial images are tailored to the specific weaknesses of downstream models, limiting their universality. Addressing this limitation to create universally crucial samples remains challenging for future work.

\section{Conclusion}
We propose Crucial-Diff, a domain-agnostic framework designed to address the challenges of data scarcity in detection and segmentation tasks. Our Crucial-Diff introduces two key modules: Weakness Aware Sample Miner module, generating crucial (hard-to-detect) samples and Scene Agnostic Feature Extractor module, allowing uniform training and synthesis. Extensive experiments validate our Crucial-Diff superiority in generating crucial samples to improve downstream task, while significantly reducing training time.


\bibliographystyle{IEEEtran}

\begin{thebibliography}{10}
\providecommand{\url}[1]{#1}
\csname url@samestyle\endcsname
\providecommand{\newblock}{\relax}
\providecommand{\bibinfo}[2]{#2}
\providecommand{\BIBentrySTDinterwordspacing}{\spaceskip=0pt\relax}
\providecommand{\BIBentryALTinterwordstretchfactor}{4}
\providecommand{\BIBentryALTinterwordspacing}{\spaceskip=\fontdimen2\font plus
\BIBentryALTinterwordstretchfactor\fontdimen3\font minus \fontdimen4\font\relax}
\providecommand{\BIBforeignlanguage}[2]{{%
\expandafter\ifx\csname l@#1\endcsname\relax
\typeout{** WARNING: IEEEtran.bst: No hyphenation pattern has been}%
\typeout{** loaded for the language `#1'. Using the pattern for}%
\typeout{** the default language instead.}%
\else
\language=\csname l@#1\endcsname
\fi
#2}}
\providecommand{\BIBdecl}{\relax}
\BIBdecl

\bibitem{roth2022towards}
K.~Roth, L.~Pemula, J.~Zepeda, B.~Sch{\"o}lkopf, T.~Brox, and P.~Gehler, ``Towards total recall in industrial anomaly detection,'' in \emph{IEEE Conf. Comput. Vis. Pattern Recog.}, 2022.

\bibitem{yao2023explicit}
X.~Yao, R.~Li, J.~Zhang, J.~Sun, and C.~Zhang, ``Explicit boundary guided semi-push-pull contrastive learning for supervised anomaly detection,'' in \emph{IEEE Conf. Comput. Vis. Pattern Recog.}, 2023.

\bibitem{you2022unified}
Z.~You, L.~Cui, Y.~Shen, K.~Yang, X.~Lu, Y.~Zheng, and X.~Le, ``A unified model for multi-class anomaly detection,'' in \emph{Adv. Neural Inform. Process. Syst.}, 2022.

\bibitem{liang2023omni}
Y.~Liang, J.~Zhang, S.~Zhao, R.~Wu, Y.~Liu, and S.~Pan, ``Omni-frequency channel-selection representations for unsupervised anomaly detection,'' \emph{IEEE Trans. Image Process.}, vol.~32, pp. 4327--4340, 2023.

\bibitem{gu2023remembering}
Z.~Gu, L.~Liu, X.~Chen, R.~Yi, J.~Zhang, Y.~Wang, C.~Wang, A.~Shu, G.~Jiang, and L.~Ma, ``Remembering normality: Memory-guided knowledge distillation for unsupervised anomaly detection,'' in \emph{Int. Conf. Comput. Vis.}, 2023.

\bibitem{liu2024cross}
B.~Liu, T.~Guo, B.~Luo, Z.~Cui, and J.~Yang, ``Cross-attention regression flow for defect detection,'' \emph{IEEE Trans. Image Process.}, vol.~33, pp. 5183--5193, 2024.

\bibitem{li2024promptad}
X.~Li, Z.~Zhang, X.~Tan, C.~Chen, Y.~Qu, Y.~Xie, and L.~Ma, ``Promptad: Learning prompts with only normal samples for few-shot anomaly detection,'' in \emph{IEEE Conf. Comput. Vis. Pattern Recog.}, 2024.

\bibitem{zhang2022lesion}
R.~Zhang, P.~Lai, X.~Wan, D.-J. Fan, F.~Gao, X.-J. Wu, and G.~Li, ``Lesion-aware dynamic kernel for polyp segmentation,'' in \emph{Med. Image Comput. Comput. Assist. Interv.}, 2022.

\bibitem{lu2024anomaly}
S.~Lu, W.~Zhang, H.~Zhao, H.~Liu, N.~Wang, and H.~Li, ``Anomaly detection for medical images using heterogeneous auto-encoder,'' \emph{IEEE Trans. Image Process.}, vol.~33, pp. 2770--2782, 2024.

\bibitem{liu2024multimodal}
H.~Liu, Z.~Ni, D.~Nie, D.~Shen, J.~Wang, and Z.~Tang, ``Multimodal brain tumor segmentation boosted by monomodal normal brain images,'' \emph{IEEE Trans. Image Process.}, vol.~33, pp. 1199--1210, 2024.

\bibitem{li2024single}
H.~Li, D.~Liu, Y.~Zeng, S.~Liu, T.~Gan, N.~Rao, J.~Yang, and B.~Zeng, ``Single-image-based deep learning for segmentation of early esophageal cancer lesions,'' \emph{IEEE Trans. Image Process.}, vol.~33, p. 2676–2688, 2024.

\bibitem{qiu2024learn}
K.~Qiu, Z.~Zhou, and Y.~Guo, ``Learn from zoom: Decoupled supervised contrastive learning for wce image classification,'' in \emph{ICASSP}, 2024.

\bibitem{li2025srconvnet}
F.~Li, R.~Cong, J.~Wu, H.~Bai, M.~Wang, and Y.~Zhao, ``Srconvnet: A transformer-style convnet for lightweight image super-resolution,'' \emph{Int. J. Comput. Vis.}, vol. 133, pp. 173--189, 2025.

\bibitem{cong2025reference}
R.~Cong, R.~Liao, F.~Li, R.~Sheng, H.~Bai, R.~Wan, S.~Kwong, and W.~Zhang, ``Reference-based iterative interaction with p 2-matching for stereo image super-resolution,'' \emph{IEEE Trans. Image Process.}, vol.~34, pp. 3779--3789, 2025.

\bibitem{hu2023anomalydiffusion}
T.~Hu, J.~Zhang, R.~Yi, Y.~Du, X.~Chen, L.~Liu, Y.~Wang, and C.~Wang, ``Anomalydiffusion: Few-shot anomaly image generation with diffusion model,'' in \emph{AAAI}, 2023.

\bibitem{ControlPolypNet}
V.~Sharma, A.~Kumar, D.~Jha, M.~Bhuyan, P.~K. Das, and U.~Bagci, ``Controlpolypnet: Towards controlled colon polyp synthesis for improved polyp segmentation,'' in \emph{CVPR Workshops}, 2024.

\bibitem{Guanfewshotgeneration}
G.~Gui, B.-B. Gao, J.~Liu, C.~Wang, and Y.~Wu, ``Few-shot anomaly-driven generation for anomaly classification and segmentation,'' in \emph{Eur. Conf. Comput. Vis.}, 2024.

\bibitem{du2023arsdm}
Y.~Du, Y.~Jiang, S.~Tan, X.~Wu, Q.~Dou, Z.~Li, G.~Li, and X.~Wan, ``Arsdm: colonoscopy images synthesis with adaptive refinement semantic diffusion models,'' in \emph{Med. Image Comput. Comput. Assist. Interv.}, 2023.

\bibitem{tan2021night}
X.~Tan, K.~Xu, Y.~Cao, Y.~Zhang, L.~Ma, and R.~W. Lau, ``Night-time scene parsing with a large real dataset,'' \emph{IEEE Trans. Image Process.}, vol.~30, pp. 9085--9098, 2021.

\bibitem{li2021cutpaste}
C.-L. Li, K.~Sohn, J.~Yoon, and T.~Pfister, ``Cutpaste: Self-supervised learning for anomaly detection and localization,'' in \emph{IEEE Conf. Comput. Vis. Pattern Recog.}, 2021.

\bibitem{niu2020defect}
S.~Niu, B.~Li, X.~Wang, and H.~Lin, ``Defect image sample generation with gan for improving defect recognition,'' \emph{IEEE Trans. Autom. Sci. Eng.}, vol.~17, pp. 1611--1622, 2020.

\bibitem{zhang2021defect}
G.~Zhang, K.~Cui, T.-Y. Hung, and S.~Lu, ``Defect-gan: High-fidelity defect synthesis for automated defect inspection,'' in \emph{WACV}, 2021.

\bibitem{duan2023few}
Y.~Duan, Y.~Hong, L.~Niu, and L.~Zhang, ``Few-shot defect image generation via defect-aware feature manipulation,'' in \emph{AAAI}, 2023.

\bibitem{radford2021learning}
A.~Radford, J.~W. Kim, C.~Hallacy, A.~Ramesh, G.~Goh, S.~Agarwal, G.~Sastry, A.~Askell, P.~Mishkin, J.~Clark \emph{et~al.}, ``Learning transferable visual models from natural language supervision,'' in \emph{ICML}, 2021.

\bibitem{goodfellow2014generative}
I.~Goodfellow, J.~Pouget-Abadie, M.~Mirza, B.~Xu, D.~Warde-Farley, S.~Ozair, A.~Courville, and Y.~Bengio, ``Generative adversarial nets,'' \emph{Adv. Neural Inform. Process. Syst.}, 2014.

\bibitem{kingma2013auto}
D.~P. Kingma, ``Auto-encoding variational bayes,'' in \emph{Int. Conf. Learn. Represent.}, 2013.

\bibitem{zhang2017stackgan}
H.~Zhang, T.~Xu, H.~Li, S.~Zhang, X.~Wang, X.~Huang, and D.~N. Metaxas, ``Stackgan: Text to photo-realistic image synthesis with stacked generative adversarial networks,'' in \emph{Int. Conf. Comput. Vis.}, 2017.

\bibitem{xu2018attngan}
T.~Xu, P.~Zhang, Q.~Huang, H.~Zhang, Z.~Gan, X.~Huang, and X.~He, ``Attngan: Fine-grained text to image generation with attentional generative adversarial networks,'' in \emph{IEEE Conf. Comput. Vis. Pattern Recog.}, 2018.

\bibitem{ding2021cogview}
M.~Ding, Z.~Yang, W.~Hong, W.~Zheng, C.~Zhou, D.~Yin, J.~Lin, X.~Zou, Z.~Shao, H.~Yang \emph{et~al.}, ``Cogview: Mastering text-to-image generation via transformers,'' \emph{Adv. Neural Inform. Process. Syst.}, 2021.

\bibitem{ramesh2021zero}
A.~Ramesh, M.~Pavlov, G.~Goh, S.~Gray, C.~Voss, A.~Radford, M.~Chen, and I.~Sutskever, ``Zero-shot text-to-image generation,'' in \emph{ICML}, 2021.

\bibitem{ho2020denoising}
J.~Ho, A.~Jain, and P.~Abbeel, ``Denoising diffusion probabilistic models,'' \emph{Adv. Neural Inform. Process. Syst.}, 2020.

\bibitem{song2019generative}
Y.~Song and S.~Ermon, ``Generative modeling by estimating gradients of the data distribution,'' in \emph{Adv. Neural Inform. Process. Syst.}, 2019.

\bibitem{rombach2022high}
R.~Rombach, A.~Blattmann, D.~Lorenz, P.~Esser, and B.~Ommer, ``High-resolution image synthesis with latent diffusion models,'' in \emph{IEEE Conf. Comput. Vis. Pattern Recog.}, 2022.

\bibitem{ramesh2022hierarchical}
A.~Ramesh, P.~Dhariwal, A.~Nichol, C.~Chu, and M.~Chen, ``Hierarchical text-conditional image generation with clip latents,'' \emph{arXiv preprint arXiv:2204.06125}, 2022.

\bibitem{gal2023image}
R.~Gal, Y.~Alaluf, Y.~Atzmon, O.~Patashnik, A.~H. Bermano, G.~Chechik, and D.~Cohen-Or, ``An image is worth one word: Personalizing text-to-image generation using textual inversion,'' in \emph{Int. Conf. Learn. Represent.}, 2023.

\bibitem{kumari2023multi}
N.~Kumari, B.~Zhang, R.~Zhang, E.~Shechtman, and J.-Y. Zhu, ``Multi-concept customization of text-to-image diffusion,'' in \emph{IEEE Conf. Comput. Vis. Pattern Recog.}, 2023.

\bibitem{kar2019meta}
A.~Kar, A.~Prakash, M.-Y. Liu, E.~Cameracci, J.~Yuan, M.~Rusiniak, D.~Acuna, A.~Torralba, and S.~Fidler, ``Meta-sim: Learning to generate synthetic datasets,'' in \emph{Int. Conf. Comput. Vis.}, 2019.

\bibitem{choi2019self}
J.~Choi, T.~Kim, and C.~Kim, ``Self-ensembling with gan-based data augmentation for domain adaptation in semantic segmentation,'' in \emph{Int. Conf. Comput. Vis.}, 2019.

\bibitem{wu2023datasetdm}
W.~Wu, Y.~Zhao, H.~Chen, Y.~Gu, R.~Zhao, Y.~He, H.~Zhou, M.~Z. Shou, and C.~Shen, ``Datasetdm: Synthesizing data with perception annotations using diffusion models,'' in \emph{Adv. Neural Inform. Process. Syst.}, 2023.

\bibitem{wu2023diffumask}
W.~Wu, Y.~Zhao, M.~Z. Shou, H.~Zhou, and C.~Shen, ``Diffumask: Synthesizing images with pixel-level annotations for semantic segmentation using diffusion models,'' in \emph{Int. Conf. Comput. Vis.}, 2023.

\bibitem{yi2023towards}
R.~Yi, H.~Tian, Z.~Gu, Y.-K. Lai, and P.~L. Rosin, ``Towards artistic image aesthetics assessment: a large-scale dataset and a new method,'' in \emph{IEEE Conf. Comput. Vis. Pattern Recog.}, 2023.

\bibitem{feng2024instagen}
C.~Feng, Y.~Zhong, Z.~Jie, W.~Xie, and L.~Ma, ``Instagen: Enhancing object detection by training on synthetic dataset,'' in \emph{IEEE Conf. Comput. Vis. Pattern Recog.}, 2024.

\bibitem{dai2024diffusion}
X.~Dai, Y.~Li, M.~Duan, and B.~Xiao, ``Diffusion models as strong adversaries,'' \emph{IEEE Trans. Image Process.}, vol.~33, pp. 6734--6747, 2024.

\bibitem{zavrtanik2021draem}
V.~Zavrtanik, M.~Kristan, and D.~Sko{\v{c}}aj, ``Draem-a discriminatively trained reconstruction embedding for surface anomaly detection,'' in \emph{Int. Conf. Comput. Vis.}, 2021.

\bibitem{lin2021few}
D.~Lin, Y.~Cao, W.~Zhu, and Y.~Li, ``Few-shot defect segmentation leveraging abundant defect-free training samples through normal background regularization and crop-and-paste operation,'' in \emph{Int. Conf. Multimedia and Expo}, 2021.

\bibitem{shin2018abnormal}
Y.~Shin, H.~A. Qadir, and I.~Balasingham, ``Abnormal colon polyp image synthesis using conditional adversarial networks for improved detection performance,'' \emph{IEEE Access}, vol.~6, pp. 56\,007--56\,017, 2018.

\bibitem{he2021colonoscopic}
F.~He, S.~Chen, S.~Li, L.~Zhou, H.~Zhang, H.~Peng, and X.~Huang, ``Colonoscopic image synthesis for polyp detector enhancement via gan and adversarial training,'' in \emph{ISBI}, 2021.

\bibitem{sams2022gan}
A.~Sams and H.~H. Shomee, ``Gan-based realistic gastrointestinal polyp image synthesis,'' in \emph{ISBI}, 2022.

\bibitem{sasmal2020improved}
P.~Sasmal, M.~Bhuyan, S.~Sonowal, Y.~Iwahori, and K.~Kasugai, ``Improved endoscopic polyp classification using gan generated synthetic data augmentation,'' in \emph{ASPCON}, 2020.

\bibitem{machavcek2023mask}
R.~Mach{\'a}{\v{c}}ek, L.~Mozaffari, Z.~Sepasdar, S.~Parasa, P.~Halvorsen, M.~A. Riegler, and V.~Thambawita, ``Mask-conditioned latent diffusion for generating gastrointestinal polyp images,'' in \emph{ACM Workshop on Intelligent Cross-Data Analysis and Retrieval}, 2023.

\bibitem{qiu2025noise}
K.~Qiu, Z.~Gao, Z.~Zhou, M.~Sun, and Y.~Guo, ``Noise-consistent siamese-diffusion for medical image synthesis and segmentation,'' in \emph{IEEE Conf. Comput. Vis. Pattern Recog.}, 2025.

\bibitem{zhou2023anomalyclip}
Q.~Zhou, G.~Pang, Y.~Tian, S.~He, and J.~Chen, ``Anomalyclip: Object-agnostic prompt learning for zero-shot anomaly detection,'' \emph{Int. Conf. Learn. Represent.}, 2023.

\bibitem{gu2024filo}
Z.~Gu, B.~Zhu, G.~Zhu, Y.~Chen, H.~Li, M.~Tang, and J.~Wang, ``Filo: Zero-shot anomaly detection by fine-grained description and high-quality localization,'' in \emph{ACM Int. Conf. Multimedia}, 2024.

\bibitem{wei2023elite}
Y.~Wei, Y.~Zhang, Z.~Ji, J.~Bai, L.~Zhang, and W.~Zuo, ``Elite: Encoding visual concepts into textual embeddings for customized text-to-image generation,'' in \emph{Int. Conf. Comput. Vis.}, 2023.

\bibitem{sun2024iseg}
L.~Sun, J.~Cao, J.~Xie, F.~S. Khan, and Y.~Pang, ``iseg: An iterative refinement-based framework for training-free segmentation,'' \emph{arXiv preprint arXiv:2409.03209}, 2024.

\bibitem{bergmann2019mvtec}
P.~Bergmann, M.~Fauser, D.~Sattlegger, and C.~Steger, ``Mvtec ad--a comprehensive real-world dataset for unsupervised anomaly detection,'' in \emph{IEEE Conf. Comput. Vis. Pattern Recog.}, 2019.

\bibitem{wang2024real}
C.~Wang, W.~Zhu, B.-B. Gao, Z.~Gan, J.~Zhang, Z.~Gu, S.~Qian, M.~Chen, and L.~Ma, ``Real-iad: A real-world multi-view dataset for benchmarking versatile industrial anomaly detection,'' in \emph{IEEE Conf. Comput. Vis. Pattern Recog.}, 2024.

\bibitem{ojha2021few}
U.~Ojha, Y.~Li, J.~Lu, A.~A. Efros, Y.~J. Lee, E.~Shechtman, and R.~Zhang, ``Few-shot image generation via cross-domain correspondence,'' in \emph{IEEE Conf. Comput. Vis. Pattern Recog.}, 2021.

\bibitem{salimans2016improved}
T.~Salimans, I.~Goodfellow, W.~Zaremba, V.~Cheung, A.~Radford, and X.~Chen, ``Improved techniques for training gans,'' in \emph{Adv. Neural Inform. Process. Syst.}, 2016.

\bibitem{binkowski2018demystifying}
M.~Bi{\'n}kowski, D.~J. Sutherland, M.~Arbel, and A.~Gretton, ``Demystifying mmd gans,'' \emph{Int. Conf. Learn. Represent.}, 2018.

\bibitem{silva2014toward}
J.~Silva, A.~Histace, O.~Romain, X.~Dray, and B.~Granado, ``Toward embedded detection of polyps in wce images for early diagnosis of colorectal cancer,'' \emph{Int. J. Comput. Ass. Rad.}, vol.~9, pp. 283--293, 2014.

\bibitem{bernal2015wm}
J.~Bernal, F.~J. S{\'a}nchez, G.~Fern{\'a}ndez-Esparrach, D.~Gil, C.~Rodr{\'\i}guez, and F.~Vilari{\~n}o, ``Wm-dova maps for accurate polyp highlighting in colonoscopy: Validation vs. saliency maps from physicians,'' \emph{Computerized medical imaging and graphics}, 2015.

\bibitem{tajbakhsh2015automated}
N.~Tajbakhsh, S.~R. Gurudu, and J.~Liang, ``Automated polyp detection in colonoscopy videos using shape and context information,'' \emph{IEEE Trans. Med. Imag.}, vol.~35, pp. 630--644, 2015.

\bibitem{vazquez2017benchmark}
D.~V{\'a}zquez, J.~Bernal, F.~J. S{\'a}nchez, G.~Fern{\'a}ndez-Esparrach, A.~M. L{\'o}pez, A.~Romero, M.~Drozdzal, A.~Courville \emph{et~al.}, ``A benchmark for endoluminal scene segmentation of colonoscopy images,'' \emph{J. Healthc. Eng.}, vol. 2017, 2017.

\bibitem{jha2020kvasir}
D.~Jha, P.~H. Smedsrud, M.~A. Riegler, P.~Halvorsen, T.~de~Lange, D.~Johansen, and H.~D. Johansen, ``Kvasir-seg: A segmented polyp dataset,'' in \emph{International conference on multimedia modeling}, 2020.

\bibitem{thambawita2022singan}
V.~Thambawita, P.~Salehi, S.~A. Sheshkal, S.~A. Hicks, H.~L. Hammer, S.~Parasa, T.~d. Lange, P.~Halvorsen, and M.~A. Riegler, ``Singan-seg: Synthetic training data generation for medical image segmentation,'' \emph{PloS One}, vol.~17, p. e0267976, 2022.

\bibitem{zhang2023prototypical}
H.~Zhang, Z.~Wu, Z.~Wang, Z.~Chen, and Y.-G. Jiang, ``Prototypical residual networks for anomaly detection and localization,'' in \emph{IEEE Conf. Comput. Vis. Pattern Recog.}, 2023.

\bibitem{xiao2024ctnet}
B.~Xiao, J.~Hu, W.~Li, C.-M. Pun, and X.~Bi, ``Ctnet: Contrastive transformer network for polyp segmentation,'' \emph{IEEE Trans. Cybern.}, 2024.

\bibitem{fan2020pranet}
D.-P. Fan, G.-P. Ji, T.~Zhou, G.~Chen, H.~Fu, J.~Shen, and L.~Shao, ``Pranet: Parallel reverse attention network for polyp segmentation,'' in \emph{Med. Image Comput. Comput. Assist. Interv.}, 2020.

\bibitem{Deng_2022_CVPR}
H.~Deng and X.~Li, ``Anomaly detection via reverse distillation from one-class embedding,'' in \emph{IEEE Conf. Comput. Vis. Pattern Recog.}, 2022.

\bibitem{liu2023simplenet}
Z.~Liu, Y.~Zhou, Y.~Xu, and Z.~Wang, ``Simplenet: A simple network for image anomaly detection and localization,'' in \emph{IEEE Conf. Comput. Vis. Pattern Recog.}, 2023.

\bibitem{zhang2023exploring}
J.~Zhang, X.~Chen, Y.~Wang, C.~Wang, Y.~Liu, X.~Li, M.-H. Yang, and D.~Tao, ``Exploring plain vit reconstruction for multi-class unsupervised anomaly detection,'' \emph{arXiv preprint arXiv:2312.07495}, 2023.

\bibitem{pang2021explainable}
G.~Pang, C.~Ding, C.~Shen, and A.~v.~d. Hengel, ``Explainable deep few-shot anomaly detection with deviation networks,'' \emph{arXiv preprint arXiv:2108.00462}, 2021.

\bibitem{ding2022catching}
C.~Ding, G.~Pang, and C.~Shen, ``Catching both gray and black swans: Open-set supervised anomaly detection,'' in \emph{IEEE Conf. Comput. Vis. Pattern Recog.}, 2022.

\bibitem{xie2021segformer}
E.~Xie, W.~Wang, Z.~Yu, A.~Anandkumar, J.~M. Alvarez, and P.~Luo, ``Segformer: Simple and efficient design for semantic segmentation with transformers,'' \emph{Adv. Neural Inform. Process. Syst.}, 2021.

\bibitem{dong2021polyp}
B.~Dong, W.~Wang, D.-P. Fan, J.~Li, H.~Fu, and L.~Shao, ``Polyp-pvt: Polyp segmentation with pyramid vision transformers,'' \emph{CAAI}, 2021.

\end{thebibliography}

\balance

\begin{IEEEbiographynophoto}{Siyue Yao}
received the B.S. degree from China University of Geosciences, Wuhan, China, in 2021 and the M.S. degree from King's College London, London, United Kingdom, in 2022. She is currently pursuing the Ph.D. degree with the chool of Advanced Technology, Xi'an Jiaotong-Liverpool University, Suzhou, China. Her research interests include generative model, image generation and motion generation.
\end{IEEEbiographynophoto}

\begin{IEEEbiographynophoto}{Mingjie Sun} (Member, IEEE) 
received the bachelor’s degree from Nanjing University of Aeronautics and Astronautics, Nanjing, China, in 2016, the master’s degree from
Xidian University, Xi’an, China, in 2019, and the Ph.D. degree from the University of Liverpool, Liverpool, U.K., in 2022. He is currently an Associate Professor with the School of Computer Science and Technology, Soochow University, Suzhou, China. His research directions mainly include computer vision, multimodal computing, and reinforcement learning.
\end{IEEEbiographynophoto}

\begin{IEEEbiographynophoto}{Eng Gee Lim} (Senior Member, IEEE) 
received the Ph.D. degree from the University of Northumbia, in 2002. He is currently a Professor with the Department of Electrical and Engineering, Xi'an Jiaotong-Liverpool University, Suzhou, China. His research interests include artificial intelligence, antennas, RF, and radio propagation for wireless communications and systems.
\end{IEEEbiographynophoto}

\begin{IEEEbiographynophoto}{Ran Yi} (Member, IEEE) 
is an Associate Professor with the School of Computer Science, Shanghai Jiao Tong University. She received the BEng degree and the PhD degree from Tsinghua University, China, in 2016 and 2021. Her research interests mainly fall into computer vision, computer graphics, and machine intelligence. She serves as a Program Committee Member/Reviewer for SIGGRAPH, CVPR, ICCV, IEEE Transactions on Pattern Analysis and Machine Intelligence, IEEE Transactions on Visualization and Computer Graphics, IEEE Transactions on Image Processing, and IJCV.
\end{IEEEbiographynophoto}

\begin{IEEEbiographynophoto}{Baojiang Zhong} (Senior Member, IEEE) received his B.S. degree in mathematics from Nanjing Normal University, China, in 1995, followed by an M.S. degree in mathematics and a Ph.D. degree in mechanical and electrical engineering from Nanjing University of Aeronautics and Astronautics (NUAA), China, in 1998 and 2006, respectively. From 1998 to 2009, he was a Faculty
Member of the Department of Mathematics, NUAA, and reached the rank of an Associate Professor. From 2007 to 2008, he was also a Research Scientist with the Temasek Laboratories, Nanyang Technological University, Singapore. In 2009, he joined the School of Computer Science and Technology, Soochow University, China, where he is currently a Full Professor. His research interests include computer vision, image processing,  image analysis and numerical linear algebra. He served as an Associate Editor for \textsc{IEEE Transactions on Image Processing} from 2022 to 2024, where he was recognized as an Outstanding Editorial Board Member in 2023. Currently, he serves as a Senior Area Editor for \textsc{IEEE Transactions on Image Processing} (2024-present) and as a Senior Area Editor for \textsc{IEEE Signal Processing Letters} (2024-present).
\end{IEEEbiographynophoto}

\begin{IEEEbiographynophoto}{Moncef Gabbouj} (Fellow, IEEE)
received the B.S. degree from Oklahoma State University, Stillwater, OK, USA, in 1985, and the M.S. and Ph.D. degrees from Purdue University, West Lafayette, IN, USA, in 1986 and 1989, respectively, all in electrical engineering.,He is now a Professor of Signal Processing with the Department of Computing Sciences, Tampere University, Tampere, Finland. He was an Academy of Finland Professor, from 2011 to 2015. His research interests include big data analytics, multimedia content-based analysis, indexing and retrieval, artificial intelligence, machine learning, pattern recognition, nonlinear signal and image processing and analysis, voice conversion, and video processing and coding.,Dr. Gabbouj is a member of the Academia Europaea and the Finnish Academy of Science and Letters. He is the Past Chair of the IEEE CAS TC on DSP and the Committee Member of the IEEE Fourier Award for Signal Processing. He is the Finland Site Director of the NSF IUCRC funded Center for Visual and Decision Informatics (CVDI) and leads the Artificial Intelligence Research Task Force of the Ministry of Economic Affairs and Employment funded Research Alliance on Autonomous Systems (RAAS). He served as an Associate Editor and a Guest Editor for many IEEE and international journals and a Distinguished Lecturer for IEEE CASS.
\end{IEEEbiographynophoto}

\vfill
\end{document}